\documentclass[journal]{IEEEtran}
\usepackage{xcolor}
\usepackage{enumitem}
\usepackage{amsmath}
\usepackage{amssymb}
\usepackage{subfigure}
\usepackage{color}
\usepackage{graphicx}
\usepackage{algorithm,algpseudocode,algorithmicx}
\usepackage{amsmath,latexsym,amssymb,amsthm,array,amsfonts,algorithm,algpseudocode,booktabs,graphicx,subfigure,multirow,cuted,stfloats}
\usepackage{footmisc}
\usepackage{cite}
\usepackage{units}
\usepackage{tabularx}
\usepackage{multirow}
\usepackage{bm}
\usepackage{float}
\usepackage{soul}
\newlength\savewidth

\ifCLASSINFOpdf
\else
\fi
\begin{document}
%
\title{Multi-objective Evolutionary Federated Learning}
%
%
%

\author{Hangyu~Zhu and
        Yaochu~Jin, \emph{Fellow}, \emph{IEEE}
\thanks{Hangyu Zhu and Yaochu Jin are with the Department of Computer Science, University of Surrey, Guildford, GU2 7XH, United Kingdom. Email: \{hangyu.zhu;yaochu.jin\}@surrey.ac.uk. (\textit{Corresponding author: Yaochu Jin})}
\thanks{Manuscript received December 6, 2018; revised xx, 2019.}
}

\maketitle

\begin{abstract}
Federated learning is an emerging technique used to prevent the leakage of private information. Unlike centralized learning that needs to collect data from users and store them collectively on a cloud server, federated learning makes it possible to learn a global model while the data are distributed on the users' devices. However, compared with the traditional centralized approach, the federated setting consumes considerable communication resources of the clients, which is indispensable for updating global models and prevents this technique from being widely used. In this paper, we aim to optimize the structure of the neural network models in federated learning using a multi-objective evolutionary algorithm to simultaneously minimize the communication costs and the global model test errors. A scalable method for encoding network connectivity is adapted to federated learning to enhance the efficiency in evolving deep neural networks. Experimental results on both multilayer perceptrons and convolutional neural networks indicate that the proposed optimization method is able to find optimized neural network models that can not only significantly reduce communication costs but also improve the learning performance of federated learning compared with the standard fully connected neural networks.
%
\end{abstract}


%
\IEEEpeerreviewmaketitle

\section{Introduction}
%
%
%
%
\IEEEPARstart{T}{he} usage of smart phones has dramatically increased over the last decades\cite{poushter2016smartphone}. Compared with classic PC devices, smart phones are more portable and user-friendly. Using smart phones has already become a significant part of modern people's daily life, while billions of data transferred between smart phones provide a great support for training machine learning models. However, traditional centralized machine learning requires local clients, e.g., smart phone users to upload their data directly to the central server for model training, which may cause severe private information leakages.

An emerging technology called federated learning \cite{mcmahan2016communication} was proposed recently to allow the central server to train a good global model, while maintaining the training data to be distributed on the clients' devices. Instead of sending data directly to the central server, each local client downloads the current global model from the server, updates the shared model by training its local data, and then uploads the updated global model back to the server. By avoid sharing local private data, users' privacy can be effectively protected in federated learning.

Some research has been dedicated to further protect users' privacy and security in federated learning. Bonawitz \emph{et al.} \cite{bonawitz2017practical} gives an overview of cryptographic techniques like homomorphic encryption \cite{atayero2011security} to encrypt the uploaded information before averaging. Different from traditional encryption methods, differential privacy \cite{dwork2008differential}, which is used to decrease individuals' information influences when querying specific data repository, protects privacy of deep learning by adding Gaussian noise \cite{abadi2016deep}. This privacy protection technology is also suited for federated learning \cite{shokri2015privacy,geyer2017differentially}.

Apart from privacy issues, statistical challenge is a barrier for federated optimization. Improving the shared global model in federated learning is sometimes similar to training the distributed model by data parallelism. McDonald \emph{et al.} proposed two distributed training strategies \cite{mcdonald2010distributed} for the structured perceptron like iterative error dependent mixing or uniform parameter mixing. Adjusted parameter mixing strategies like fish matrix implementation \cite{povey2014parallel} and elastic averaging stochastic gradient descent \cite{zhang2015deep} can further improve the convergence efficiency and robustness in distributed model mixture. However, the aforementioned algorithms are built under the assumption that data on each local edge is independent and identically distributed (IID), and non-IID local data distribution was not considered. To address this problem, Zhao \emph{et al.} \cite{zhao2018federated} did some experiments on highly skewed non-IID data and provided statistically divergence analysis.

Federated learning requires massive communication resources compared to the classic centralized learning. A federated averaging algorithm \cite{mcmahan2016communication} introduced by McMahan \emph{et al.} can improve communication efficiency by reducing local training mini-batch sizes or increasing local training passes to reduce communication rounds. Shokri \emph{et al.} used the method of uploading the gradients located in the particular interval clipped by some threshold values \cite{shokri2015privacy}, which is similar to the idea of structured updates introduced in \cite{konevcny2016federated}.

Another method to reduce the communication cost is to scale down the uploaded parameters by reducing the complexity of the neural network models. The early ideas of evolving artificial neural network were introduced in \cite{yao1999evolving}, where systematic neural network encoding methods were presented. However, most of them are direct encoding methods that are not easily scalable to deep neural networks having a large number of layers and connections. In order to address this issue, neuroevolution of augmenting topologies (NEAT) \cite{stanley2002evolving} and undirect graph encoding \cite{fekiavc2011review} were proposed to enhance the flexibility of neural network encoding. Although they are able to substantially improve the encoding efficiency, both NEAT and cellular graph method occupy too many computation resources. More recently, Mocanu proposed a sparse evolutionary algorithm (SET) \cite{mocanu2018scalable} to reduce the search space in optimizing deep neural networks containing a large number of connections.

To reduce the communication costs without seriously degrading the global learning accuracy, this work proposes a framework for optimizing deep neural network models in federated learning. The main contributions of the paper are as follows:
\begin{enumerate}
  \item Federated learning is formulated as a bi-objective optimization problem, where the two objectives are the minimization of the communication cost and the maximization of the global learning accuracy. This bi-objective optimization is solved by a multi-objective evolutionary algorithm.
  \item A modified SET algorithm is proposed to reduce the connections of neural networks, thereby indirectly reducing the number of model parameters to be sent to the server.
\end{enumerate}
Our experimental results indicate that the proposed algorithm can significantly reduce the complexity of the neural network models at the expense of minor performance degradation of the global model, thereby reducing the server-client communication.

The rest of this paper is organized as follows. Section II introduces the related background. A detailed description of the proposed algorithms are given in Section III. In Section IV, the experimental results are presented and discussed. Finally, the paper is concluded in Section V.

\section{Preliminaries}
In this section, we briefly review the basics of multilayer perceptron and convolutional neural networks, federated learning, and evolutionary optimization of neural networks.

\subsection{Multilayer perceptron neural networks}
Multilayer perceptrons (MLP) \cite{pham1970neural} are the most commonly used feedforward artificial neural networks containing at least three layers: the input layer, one hidden layer and the output layer. Typically, nodes or neurons located between each layer of the MLP are fully connected without internal loops and use activation functions for the purpose of non-linear projection and feature extraction upon outputs from the previous layer.

\begin{figure}
\centering
\includegraphics[height=6cm, width=5cm]{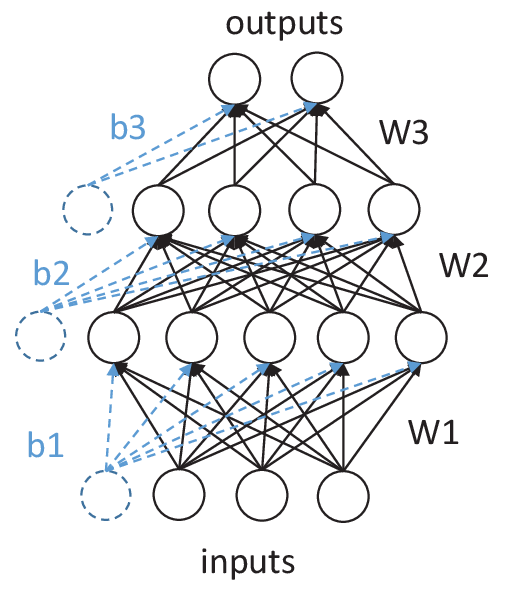}
\caption{This multilayer perceptron neural network contains an input layer, two hidden layers, and an output layer. Solid circles represent neurons while dashed ones represent biases.}
\label{Fig_1}
\end{figure}

Fig. 1 shows an illustrative example of a fully connected multilayer perceptron. Circles in solid lines in the figure represent 'neurons' and circles in dashed line are called 'biases'. In the feed-forward propagation of a fully connected neural network, each node or neuron receives a weighted sum of the inputs of all preceding neurons plus a bias value as its input. Then the output of this neuron is computed by a nonlinear activation function $ \sigma  $ as follows:
\begin{equation}
\begin{split}
{y_{neuron}}  = \sigma (\sum\limits_{i = 1}^N {{x_i} \cdot {w_i} + b) = \sigma ({x^T}w + b)} \
\label{eq1}
\end{split}
\end{equation}
When the feed-forward propagation passes through one or more hidden layers to the output layer, a predicted target $\hat{y}$ is achieved to compute the loss function $ \ell (\hat{y},y)\ $, which is typically the difference between the desired output $y$ and predicted output $\hat{y}$. If we use $\theta$ to replace both weights and biases, the loss function can be reformulated as $ \ell (\theta ) $ and then the neural network tries to optimize the trainable parameter $\theta$ by minimizing the loss $ \ell (\theta )$.
\begin{equation}
\begin{split}
\mathop {\min } \limits_\theta \ \ell (\theta ) = \frac{1}{N}\sum\limits_i {\ell (\theta ,{x_i})}\quad x_i \in \left\{ {{x_1},{x_2}...,{x_N}} \right\}
\label{eq2}
\end{split}
\end{equation}
where $ {x_i}\ $ is the $i$-th training sample (can be a vector) and $ N\ $ is the size of training data. The objective is to find a specific parameter $ \theta  $ to minimize the \emph{expected} loss through $N$ data samples.

Gradient descent (GD) is commonly used to train neural networks in the back-propagation by computing the partial derivative of a loss function $ \ell (\theta )$ over the whole $N$ data samples with respect to each element in $\theta$. However, this approach takes a very long time to compute the gradient in each iteration if the total number of input samples is very large. The stochastic gradient descent (SGD) algorithm is at another extreme compared to GD -- it only randomly chooses one training sample per iteration, which however, may cause instability in training. To strike a balance between computation efficiency and training stability, mini-batch stochastic gradient descent (mini-batch SGD) is proposed to select a randomly chosen mini-batch size of the training data for gradient computation in each training iteration:
\begin{equation}
\begin{split}
{g_t} = \frac{1}{n}{\nabla _\theta }\ell (\theta ,{x_{i:i + n}})\ \\
{\theta _{t + 1}} = {\theta _t} - \eta {g_t}\
\label{eq3}
\end{split}
\end{equation}
where $n$ is the size of mini-batch, $\eta$ is the learning rate, and ${g_t}$ is the \emph{average} gradient over data samples $x_{i:i + n}$ with respect to the elements in $\theta _t$ in the $t$-th iteration. The training of the neural network is to update the parameter $\theta$ by iteratively subtracting $\eta {g_t}$ from the current model parameter $\theta _t$.

\subsection{Convolutional neural networks}
Convolutional neural networks (CNNs) \cite{lecun1995convolutional} are well suited for dealing with very high dimensional inputs and have shown consistently better performances than MLPs for image classification. CNNs share a similar topological architecture with MLPs, but several variations are made to CNNs based on the structure of MLPs.

A CNN generally has three kinds of layers: convolutional layers, pooling layers and fully connected layers. The convolutional layer consists of numerous kernel filters that can be recognized as an array of square block neurons, where the real number inside each square neuron is equivalent to the connection in the MLP. The convolutional layer does the 'convolution' operations on the previous layer, where the kernel filters can be seen as training weights. The CNN can be mathematically described as follows:
\begin{equation}
\begin{split}
\mathop y\nolimits_{ij}^l  = \sigma (\sum\limits_{a = 0}^{n - 1} {\sum\limits_{b = 0}^{n - 1} {{f_{ab}}x_{(i + a)(j + b)}^{l - 1}} )}\
\label{eq4}
\end{split}
\end{equation}
where ${f_{ab}}$ is a $ n \times n\ $ kernel filter, $l$ is the layer number, $ {x^{l - 1}}\ $ is the input of the convolutional layer, $ y_{ij}^l $ is the output of the convolutional layer, and $\sigma$ is the activation function. Specifically, we use the rectified linear unit (relu) as our hidden neuron's activation function to relieve the effect of gradient vanishing \cite{hochreiter1998vanishing} and softmax function in the output nodes for multi class classification tasks. Formulas of relu and softmax function are shown below:
\begin{equation}
\begin{split}
{\sigma _{relu}}(z) = max(0,z)\ \\
{\sigma _{soft\max }}({z_i}) = \frac{{\exp ({z_i})}}{{\sum\limits_{i = 1}^{C} {\exp ({z_i})} }}\
\label{eq4}
\end{split}
\end{equation}
where $z$ is the output of the previous layer and $C$ is the total number of label classes we need to classify.

A pooling layer can be added in the CNN after several convolutional layers \textcolor{red}{for specific feature extractions of hidden representations}. For instance, a dimension of $ m \times m $ Max pooling window is generally created to extract the maximum luminance value of pixels within the corresponding Max pooling window area for further enhancing the representation features of filtered images from the previous convolutional layer. Besides the Max pooling operation, Average pooling method is also commonly used by instead averaging feature values among the window area.

The fully connected layer is applied at the back of the CNN. It is exactly the same as a traditional neuron layer in the MLP, with its input being the flattened image pixels from the output of its preceding layer. The main purpose of this layer is to classify the extracted features from the previous layers in the CNN into various classes.

The aforementioned mini-batch GD is also applicable to the CNN. It should be noticed that we only calculate the partial derivative of weights in the convolutional and fully connected layers and do nothing with pooling layers when performing the back-propagation optimization on CNNs. This is because the pooling operation does not contain any trainable parameters with respect to the derivatives of the back-propagation.

\subsection{Federated learning}
Federated learning \cite{mcmahan2017federated} is an emerging decentralized privacy-protection training technology that enables client edges to learn a shared global model without uploading their private local data to a central server. In each training round, a local device downloads a shared model from the global server cloud, trains the downloaded model over the individuals' local data and then sends the updated weights or gradients back to the server. On the server, the uploaded models from the clients are aggregated to obtain a new global model. Compared with the traditional centralized learning, federated learning has the following unique features:
\begin{enumerate}
  \item The training data are distributed on the local edges, which is not available to the global server cloud. However, the learned model is shared between the server and all clients.
  \item Model training occurs on each local device instead of on the server. The server aggregates the local models uploaded from the clients to obtain a shared global model and send the global model back to the clients.
  \item Federated learning has a much higher requirement on local computation powers and communication resources than the traditional centralized learning.
\end{enumerate}
Similar to the learning algorithm of the multilayer perceptron neural network, federated learning aims to minimize the loss function $\ell (\theta )$ but in a distributed scheme:
\begin{equation}
\begin{split}
\mathop {\min }\limits_\theta \ \ell (\theta ) = \sum\limits_{k = 1}^K {\frac{{{n_k}}}{n}{L_k}(} \theta )\quad \textrm{where}\quad {L_k}(\theta ) = \frac{1}{{{n_k}}}\sum\limits_{i \in {P_k}} {{\ell _i}(\theta )} \\
\end{split}
\label{eq5}
\end{equation}
where $k$ is the index of $K$ total clients, ${L_k}(\theta )$ is the loss function of $k$-th local client, ${n_k}$ equals to the local data size, and ${P_k}$ is the set of data indexes whose length is $ {n_k} $, i.e., $ {n_k} = |{P_k}|\ $. Optimizing the loss function $\ell (\theta )$ in federated learning is equivalent to minimizing the weighted average of local loss function ${L_k}(\theta )$.

\begin{figure}
\includegraphics[height=7cm, width=8cm]{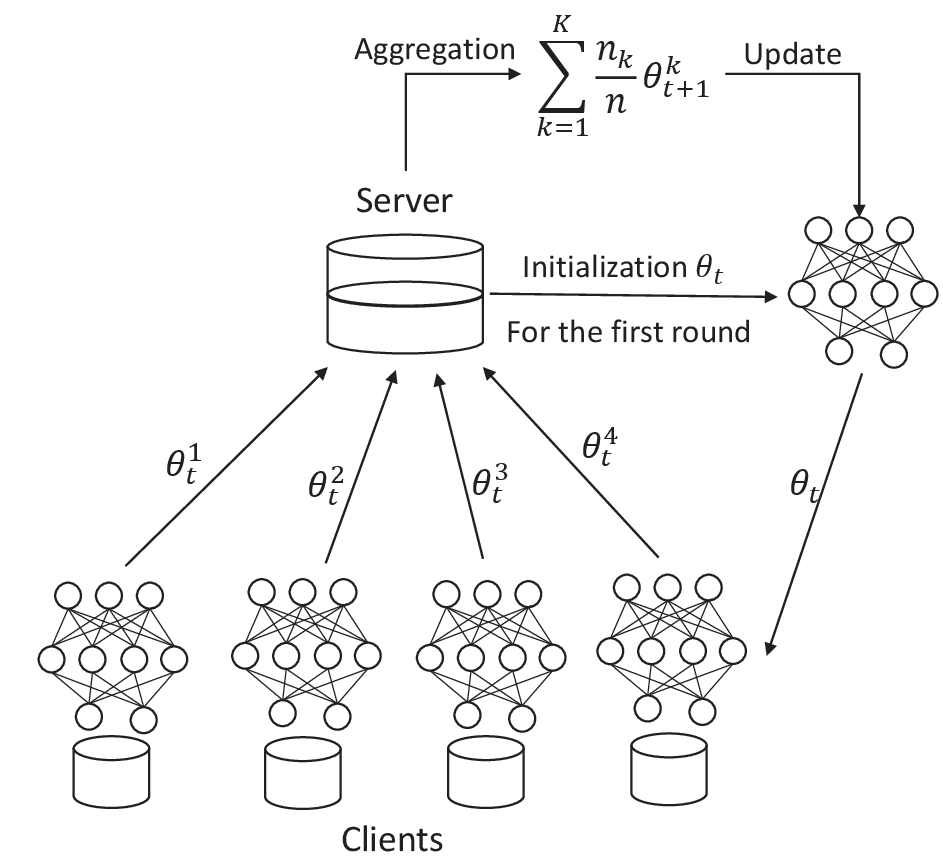}
\centering
\caption{Flowchart of federated learning. $ \theta  $ is the model parameters transferred between the server and clients, $ {n_k}\ $is the data size of client $k$, $K$ is the total number of clients and $t$ is the communication round in federated learning. We just initialize global model parameters randomly at the beginning of the communication round and use updated model parameters afterwards}
\label{Fig_2}
\end{figure}
The procedure of federated learning is shown in Fig. 2, where each client receives the parameters $ {\theta _t}\ $ of the global model from the central server and then trains their individual local models using their own data. After local training, each local device sends their trained local parameters (for instance $ \theta _t^1 $) to the server to be aggregated to get an updated global model $ {\theta _{t + 1}} $ to be used for the next iteration's training. The subscript $t$ denotes the time sequences or so called communication rounds in federated learning.

The federated averaging (FedAvg) algorithm \cite{mcmahan2016communication} can effectively reduce communication rounds by simultaneously increasing local training epochs and decreasing local mini-batch sizes in federated stochastic gradient descent algorithm (FedSGD) \cite{mcmahan2016communication}. The pseudo code of FedAvg is presented in \textbf{Algorithm 1}, where $\theta ^k$ are the model parameters of the $k$-th client.
\begin{algorithm}[htbp]\footnotesize{
\caption{FederatedAveraging. $K$ indicates the total numbers of clients; $B$ is size of mini batch, $ E\ $ is equal to training iterations and $ \eta \ $ is the learning rate} \algblock{Begin}{End}
\label{Algorithm 1}
\begin{algorithmic}[1]
\State \textbf{Server: }
\State Initialize $ {\theta _t}\ $
\For {each communication round $ t = 1,2,...\ $}
\State Select $ m = C \times K\ $ clients, $ C \in (0,1)\ $ clients
\State Download $ {\theta _t}\ $ to each client $k$
\For {each client $ k \in m\ $}
\State Wait \textbf{Client $k$ } for synchronization
\State $ {\theta _t} = \sum\limits_{k = 1}^m {\frac{{{n_k}}}{n}{\theta ^k}} \ $
\EndFor
\EndFor
\State \textbf{Client $k$: }
\State $ {\theta ^k} = {\theta _t}\ $
\For {each iteration from 1 to $E$}
\For {batch $ b \in B\ $}
\State $ {\theta ^k} = {\theta ^k} - \eta \nabla {L_k}({\theta ^k},b)\ $
\EndFor
\EndFor
\State return ${\theta ^k}$ to server
\end{algorithmic}}
\end{algorithm}

In the \textbf{Algorithm 1}, $n$ is the size of the whole data, and the global model parameter $\theta _t$ over $t$-th communication round is calculated by a weighted average of $\theta ^k$ from each client $k$. The client selection parameter $C$ is a random number between 0 to 1 determining the total fraction of {$ C \times K $} clients allowed to update the shared global model.

The number of clients participating in federated learning may heavily affect the training performance, if the data on each clients do not cover the distribution of the overall data, which is very likely to happen in federated learning. As already found in \cite{mcmahan2016communication}, selecting more clients for training can speed up the convergence of the global model and enhance its performance, if the data on the participating clients in each communication round cannot cover the overall data distribution. More recently, it was theoretically shown \cite{zhao2018federated} that the global weight convergence can be affected by the probability differences between data distributed on client $k$ and the whole data population, i.e., $ \sum\limits_{i = 1}^L {||{p^k}(y = i) - p(y = i)||} \ $, where $L$ represents the total label classes, $ {p^k}(y = i) $ is the probability of data occurrence corresponding to the label $i$ for client $k$ and  $ p(y = i) $ is that for the whole data population, respectively. Therefore, data distribution discrepancy on the client side is a root cause of the weight divergence and clients with non-IID data is harder to train than those with IID data. Unfortunately, selecting the right number of participating clients is challenging in federated learning since the class balance, data distribution and the amount of the data may vary a lot from client to client, and also over time.

It should be mentioned that different observations have been made in other contexts of distributed learning where the data can be proactively divided over clients. For example, it was suggested in \cite{mcdonald2010distributed} that increasing the number of client shards may slow down the convergence of the weights in the IID environment. This happens because if the data distributed on the local devices, which are selected to communicate with the central server, can cover the whole data population, the client or replicas that has a larger data size converges to its optimum more quickly. Thus, the larger the number of client shards is, the smaller the expected amount of data that can be allocated to each client will be, if the whole data size is fixed. On the contrary, if the selected clients only hold a fraction of the whole training data, information deficiency of clients' data may cause negative effect on convergence performance.

\subsection{The elitist non-dominated sorting genetic algorithm}
We adopt the elitist non-dominated sorting genetic algorithm  (NSGA-II) \cite{deb2002fast}, a widely used multi-objective evolutionary algorithm, to optimize the connectivity and hyper parameters of the neural network to simultaneously minimize the communication costs and maximize the global learning accuracy. NSGA-II is able to achieve a set of diverse Pareto optimal solutions by comparing the dominance relationships between the solutions in the population and a crowding distance calculated according to the distance between two neighbouring solutions. The main procedure of NSGA-II can be summarized as follows:

\begin{enumerate}
  \item \textbf{Step 1:} Randomly generate a parent population $ {P_t} $ of a size $N$ for the first generation.
  \item \textbf{Step 2:} Create an offspring population $ {Q_t} $ of the same size as the parent population $ {P_t} $ by using crossover and mutation operators on $ {P_t} $. Merge $ {P_t} $ and $ {Q_t} $ into a combined population $ {R_t} $, where $ {R_t} = {P_t} \cup {Q_t} $ has a size of $2N$.
  \item \textbf{Step 3:} Perform non-dominated sorting to sort the combined population $ {R_t} $ into a number of non-dominated fronts according to their dominance relationships. Thus, solutions in the same front are non-dominated with each other. Once the combined population is sorted, calculate the crowding distance for each individual in the same non-dominated front based on the distance to its neighboring solutions. Note that the solutions at the ends of each non-dominated front are an infinite number so that they are always prioritized in selection.
  \item \textbf{Step 4:} Generate the parent population for the next generation $ {P_{t + 1}} $ by selecting $N$ better solutions front by front from the sorted combined population $ {R_t} $. If the number of solutions located in the selected last non-dominated front is larger than that of solutions remains to be selected for $ {P_{t + 1}} $, the individuals with a larger crowding distance will be selected to promote the diversity of the population.
  \item \textbf{Step 5:} Go to \textbf{step 2} and repeat the whole procedure until a stop criterion is met.
\end{enumerate}

NSGA-II is a powerful and robust multi-objective evolutionary algorithm for problems having two or three objectives. More recent evolutionary algorithms can be adopted if the number of objectives is larger than three, e.g., the evolutionary many-objective optimization algorithm using reference-point based non-dominated sorting approach \cite{deb2013}, the knee-driven evolutionary algorithm for many-objective optimization \cite{zhang2015knee}, and the reference vector guided evolutionary for many-objective optimization \cite{cheng2016}. Note also that computationally more efficient non-dominated sorting algorithms can be used when the population size is large \cite{zhang2015ESN} or when the number of objectives is large \cite{zhang2018}.

The use of the NSGA-II to optimize the neural network model in federated learning will certainly increase the computational complexity of the algorithm. In NSGA-II, the fast non-dominated sorting operation has a computational complexity of $O(m{N^2})$ \cite{deb2000fast}, where $m$ is the number of objectives and $N$ is the number of populations. The computation complexity of crowding distance calculation is $ O(mN\log N)\ $ in the worst case, when all the solutions are located in one non-dominated front. Note, however, that in practice, the majority of the computational complexity mainly comes from the large number of time-consuming evaluations of the objective functions. For example, each evaluation of the objective functions in evolutionary optimization of the neural networks requires the training of the model, which can be computationally intensive if the amount of data is large. To address this issue, surrogate-assisted evolutionary optimization \cite{jin2011surrogate,jin2018datadriven} or Bayesian optimization \cite{Bobak2016} are helpful to reduce the computation cost.

\section{Proposed Algorithm}
In this section, we at first introduce the modified sparse evolutionary training algorithm. Then we formulate federated learning as a bi-objective optimization problem. This is followed by a description of the encoding scheme adopted by the evolutionary algorithm. Finally, the overall framework is presented.

\subsection{The modified sparse evolutionary training algorithm}
In evolutionary optimization of the structure of neural networks, the encoding scheme used by the evolutionary algorithm significantly affects the optimization efficiency. Direct binary encoding such as the one introduced in \cite{jin2008pareto} needs a large connection matrix to represent the structure of a neural network, which is not scalable to neural networks containing multiple hidden layers  and a large number of neurons. In order to enhance the scalability in evolving deep neural networks, we propose a modified sparse evolutionary training (SET) \cite{mocanu2018scalable} method to simultaneously improve the scalability and flexibility in evolving neural networks.

SET is different from typical methods for evolving the structure of neural networks. It does not directly encode the neural network and perform selection, crossover and mutation as done in genetic algorithms \cite{whitley1994genetic}. Instead, SET starts from an initial Erdos Rényi random graph \cite{erdos1960evolution} that determines the connectivity between every two neighboring layers of the neural network. The connection probability between two layers is described as follows in Eq. (7):
\begin{equation}
\begin{split}
p(W_{ij}^k) = \frac{{\varepsilon ({n^k} + {n^{k - 1}})}}{{{n^k}{n^{k - 1}}}}\ \\
{n^W} = {n^k}{n^{k - 1}}p(W_{ij}^k)\
\label{eq7}
\end{split}
\end{equation}
where $ {n^k}\ $ and $ {n^{k-1}}\ $ are the number of neurons in layer $k$ and $k-1$, respectively, $ {W_{ij}^k} $ is the sparse weight matrix between the two layers, $ \varepsilon $ is a SET parameter that controls connection sparsity, and ${n^W}$ is the total number of connections between the two layers. It is easy to find that the connection probability would become significantly lower, if $ \varepsilon  \ll {n^k}\ $ and $ \varepsilon  \ll {n^{k - 1}}\ $.

Since the randomly initialized graph may not be suited for learning a particular data, Mocanu \emph{et al.} suggest removing a fraction $ \xi  $ of the weights with the smallest update during each training epoch, which can be seen as the selection operation of an evolutionary algorithm. However, removing the least important weights may cause fluctuation when minimizing the loss function using the mini-batch SGD algorithm and this phenomenon turns out to be extremely severe in federated learning. To address this issue, we modify the operator by conducting the removal operation at the \emph{last} training epoch only. Pseudo code of the modified SET is listed in \textbf{Algorithm 2}.
\begin{algorithm}[htbp]\footnotesize{
\caption{Modified sparse evolutionary training algorithm} \algblock{Begin}{End}
\label{Algorithm 2}
\begin{algorithmic}[1]
\State Set $\varepsilon$ and $ \xi \ $
\For {each fully-connected layer of the neural network}
\State Replace weight matrices by Erdos Rényi random graphs given by $\varepsilon$ in Eq. (7)
\EndFor
\State Initialize weights
\State Start training
\For {each training epoch}
\State Training and updating corresponding weights
\EndFor
\For {each weight matrix}
\State Remove a fraction $ \xi \ $ of the smallest $ |weights|\ $
\EndFor
\end{algorithmic}}
\end{algorithm}
By implementing the modified SET algorithm, a sparsely connected neural network can be evolved, resulting in much fewer parameters to be downloaded or uploaded, thereby reducing the communication cost in federated learning.

\subsection{The objective functions and encoding of the neural networks}
We reformulate federated learning as a two objective optimization problem \cite{deb2014multi}. One objective is the global model test error $ {E_t} $ and the other is the model complexity $ {\Omega _t} $ over the $t$-th communication round. To minimize these two objectives, we evolve both the hyper parameters as well as the connectivity of the neural network models. The hyper parameters include the number of hidden layers, the number of neurons in each hidden layer, and the learning rate $\eta$ of the mini-batch SGD algorithm. The connectivity of the neural network is represented by the modified SET algorithm described in \textbf{Algorithm 2}, which consists of two parameters, namely, $\varepsilon$ in Eq. (7), an integer, and the fraction of weights to be removed, $\xi$, a real number between 0 and 1.

Consequently, we have two types of decision variables to be encoded in the chromosome of the evolutionary algorithm, i.e., real numbers and integers. Here, all integers are encoded  using binary coding and all real-valued parameters are real-encoded. For instance, the number of hidden layers and the number of nodes in each layer should be converted into binary numbers, while the real-valued parameters like learning rate and SET variables remain to be real values. Fig. 3 provides an example of an encoded individual and the corresponding MLP neural network, where $\xi=0.3$, the learning rate $\eta=0.1$, and $\varepsilon=20$. In addition, the network has two hidden layers, each containing five and four neurons, respectively.

The encoding of the CNN is slightly different, mainly because a CNN contains a number of convolutionary layers followed by a number of fully connected classification layers. Integers like the number of convolutional layers and the number of output channels for each convolutional layer are encoded using binary numbers. We just choose a value randomly between integer $3$ and $5$ for convenience for the kernel size. Refer to Fig. \ref{Fig_CNN} for an illustrative example.
\begin{figure}
\includegraphics[height=7cm, width=8cm]{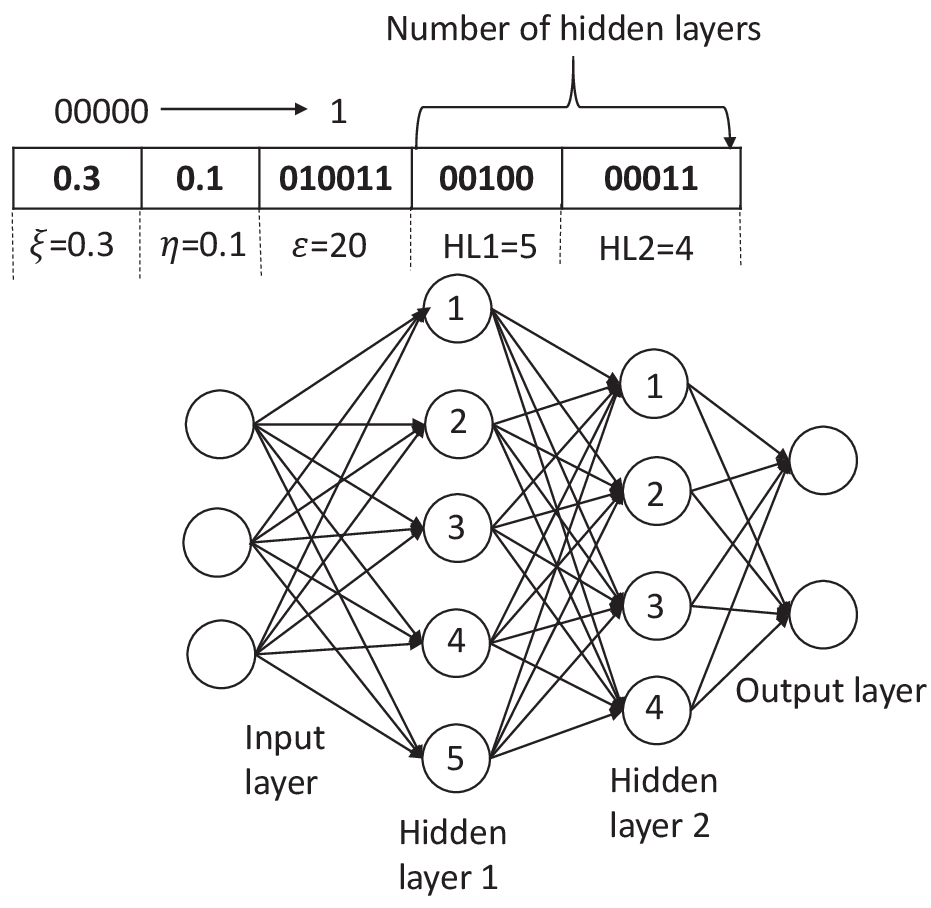}
\caption{A neural network and its chromosome. Note that when we decode the number of neurons, each variable will be increased by one to make sure that there is at least one neuron in a hidden layer. In the figure, HL1 and HL2 denote that the neural network has two hidden layers containing 5 and 4 neurons, respectively.}
\label{Fig_3}
\end{figure}

\begin{figure}
\centering
\includegraphics[height=6cm, width=8cm]{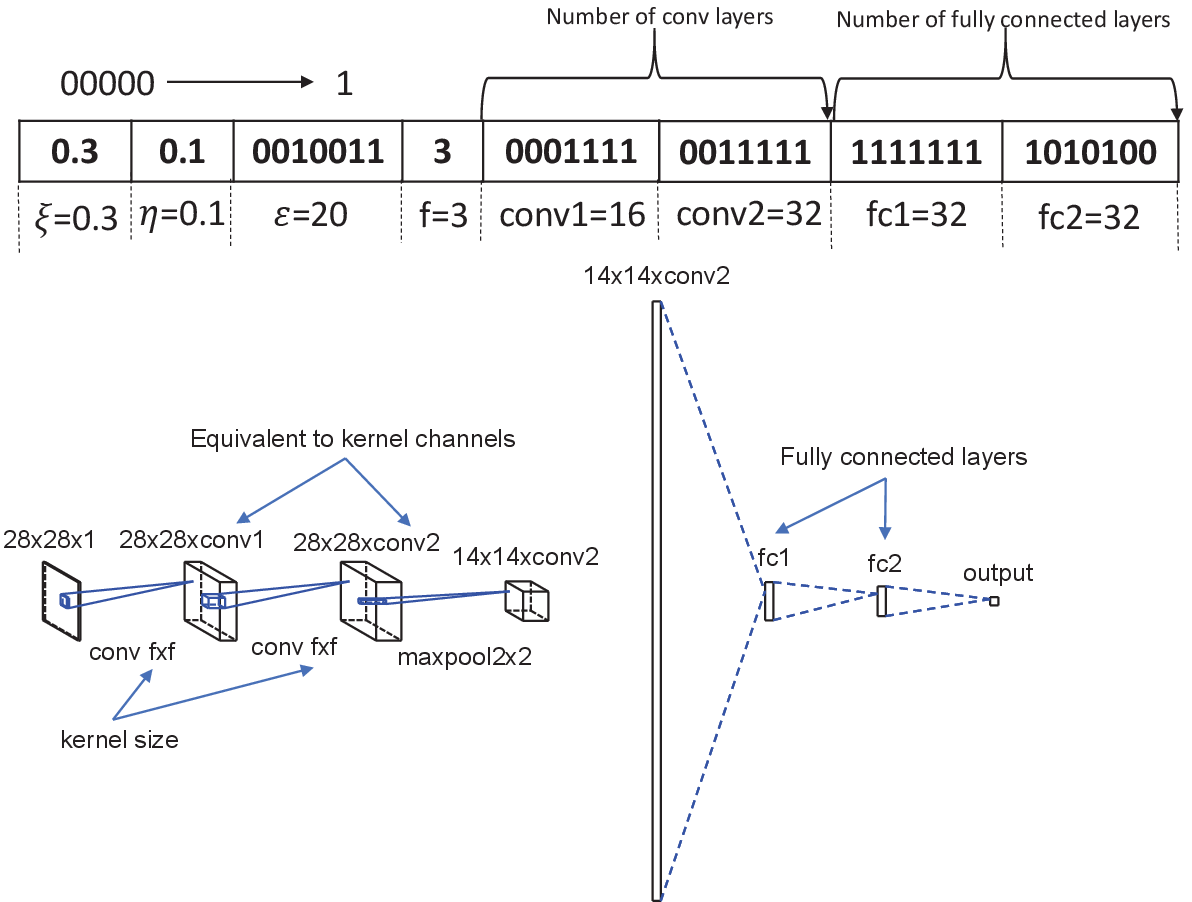}
\caption{An illustrative example of an individual encoding a convolutional neural network. Note that the minimum number of neurons in each hidden layer is 1. Two chromosomes conv1 and conv2 represent 16 and 32 filter channels, respectively, and fc1 and fc2 represent 32 and 32 neurons, respectively, in the fully connected layers. $f$ is the size of convolution kernel or filter. The padding type is set to be 'same', so the size of featured images for each convolutional layer output remains the same before the Max pooling operation.}
\label{Fig_CNN}
\end{figure}

After generating a sparsely connected neural network model, we use the FedAvg algorithm to train the network and calculate the test accuracy $ {A_t} $ within a certain number of communication rounds $t$. This global test accuracy will be used to calculate the test error $ {E_t} $ of the global model, which is one of the objectives of the bi-objective optimization problem. The model complexity $ {\Omega _t} $, the other objective, can be measured by averaging the number of weights uploaded from all clients in the $t$-th communication round:
\begin{equation}
\begin{split}
{E_t} = 1 - {A_t}\ \\
{\Omega _t} = \sum\limits_{k = 1}^K {{\Omega _k}} /K\
\label{eq8}
\end{split}
\end{equation}
where $K$ is the total number of clients and ${\Omega _i}$ indicates the number of parameters of the $k$-th client model.

\subsection{The modified SET federated averaging algorithm}
As mentioned above, the learning performance is evaluated by calculating the test error of the federated global model trained by the FedAvg algorithm (\textbf{Algorithm 1}). The modified SET algorithm is then integrated with the FedAvg algorithm to reduce the connectivity of the shared neural network model. The modified SET FedAvg optimization is described in \textbf{Algorithm 3}.
\begin{algorithm}[htbp]\footnotesize{
\caption{The modified SET FedAvg optimization. $K$ indicates the total numbers of clients, $k$ represents the $k$-th local client, $B$ is the local mini-batch size, $E$ is the number of local training iterations, $ \eta \ $ is the learning rate,
$ \Omega \ $ represents the number of connections, $\varepsilon$ and $ \xi \ $ are both SET parameters introduced in \textbf{Algorithm 3}}\algblock{Begin}{End}
\label{Algorithm 3}
\begin{algorithmic}[1]
\For {each population $ {i} \in {R}\ $}
\State Globally initialize $ {\theta _t^i}\ $ with a Erdos Rényi topology given by $\varepsilon$ and \textcolor{red}{$ \xi \ $} in Eq. (7)
\For {each communication round $ t = 1,2,...\ $}
\State Select $ m = C \times K\ $ clients, $ C \in (0,1)\ $ clients
\State $ {\Omega _t} = 0\ $
\For {each client $ k \in {m}\ $}
\For {each local epoch $e$ from 1 to $E$}
\For {batch $ b \in B\ $}
\State $ \theta _e^k = \theta _t^i - \eta \nabla \ell (\theta _t^i;b)\ $
\EndFor
\State remove a fraction of $ \xi \ $ smallest values in $ \theta ^k $
\EndFor
\State $ \theta _{t + 1}^i = \theta _t^i + \frac{{{n_k}}}{n}{\theta ^k}\ $
\State $ {\Omega ^k} = f({\theta ^k})\ $ (calculate the number of weight parameters)
\State $ {\Omega _t} = {\Omega _t} + \frac{{{n_k}}}{n}{\Omega ^k}\ $
\EndFor
\EndFor
\State Evaluate test accuracy through $ {\theta ^i} $ and test dataset
\State Calculate test error as objective one $ f_i^1\ $
\State Set $ \Omega _t $ as objective two $ f_i^2\ $
\EndFor
\State return $ f^1 $ and $ f^2 $
\end{algorithmic}}
\end{algorithm}

In the algorithm, $i$ is one solution that represents a particular neural network model with a modified SET topology as a global model used in FedAvg and $R$ is the population size. Once the hyper parameters and the connectivity of the neural network are determined by the evolutionary algorithm, the weights will be trained using the mini-batch SGD and the global model will be updated. This process repeats for a certain number of communication rounds before the two objectives can be calculated.

\subsection{Multi-objective evolutionary optimization}
The bi-objective optimization of federated learning can be solved using any multi-objective evolutionary algorithms. Here, we employ the popular NSGA-II for achieving a set of Pareto optimal solutions. A diagram of the overall algorithm is plotted in Fig. 5, and the pseudo code is summarized in \textbf{Algorithm 4}.

\begin{figure}
\includegraphics[height=6cm, width=8cm]{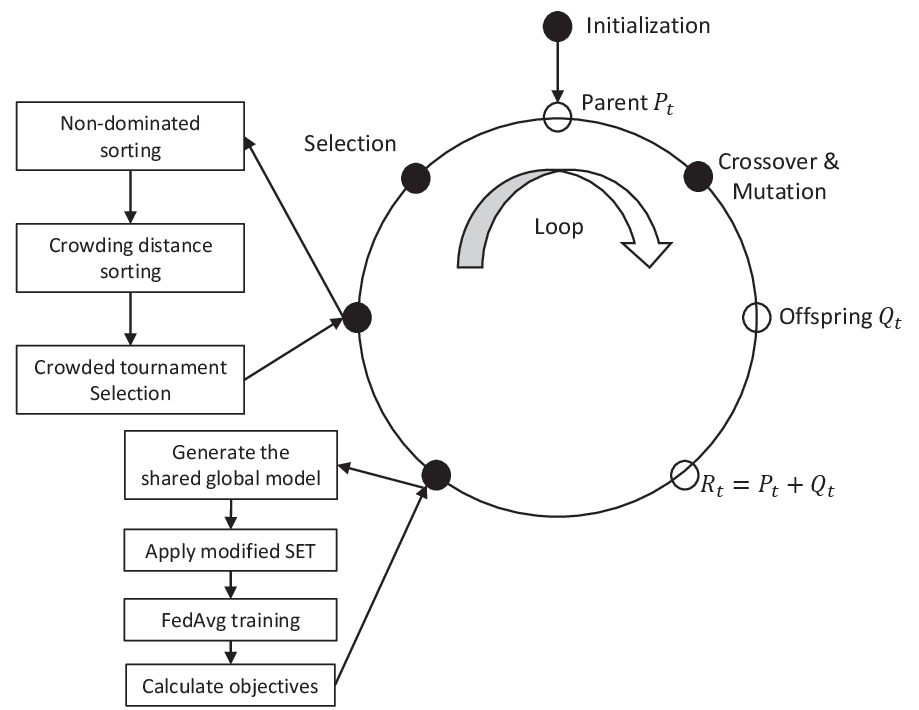}
\caption{A framework for multi-objective optimization of federated learning using NSGA-II.}
\label{Fig_4}
\end{figure}

\begin{algorithm}[htbp]\footnotesize{
\caption{Multi-objective evolutionary optimization} \algblock{Begin}{End}
\label{Algorithm 4}
\begin{algorithmic}[1]
\State Randomly generate parent solutions $ {P_t} $ where $ |{P_t}| = M $
\For {each generation $ t = 1,2,...\ $}
\State Generate offspring $ |{Q_t}| = M $ through crossover and mutation
\State $ {R_t} = {P_t} + {Q_t}\ $
\State Evaluate $ f_t^1\ $ and $ f_t^2\ $ by Algorithm 5
\State $ f \leftarrow (f_t^1,f_t^2)\ $
\For {each solution in $ {R_t}\ $}
\State Do non-dominated sorting and calculate crowding distance on $f$
\State Select high-ranking solutions from $ {R_t} $
\State Let $ {P_t} = {R_t}\ $
\EndFor
\EndFor
\end{algorithmic}}
\end{algorithm}

NSGA-II begins with the initialization of the population of size $M$ where the binary and real-valued chromosomes are randomly initialized, which is the parent population at the first generation. Two parents are selected using the tournament selection to create two offspring by applying one-point crossover and flip mutation on the binary chromosome and the simulated binary crossover (SBX) and polynomial mutation \cite{agrawal1995simulated} on the real-valued chromosome. This process repeats until $M$ offspring are generated.

We then calculate the two objectives of  each individual in the offspring population. After that, the parent and offspring populations are combined and sorted according to the non-dominance relationship and crowding distance. Finally,  $M$ high-ranking individuals from the combined population are selected as the parent of the next generation.

We repeat above procedure for several generations to generate a set of non-dominated solutions.

\section{Experimental Results}
Two experiments are designed to examine the performance of the proposed multi-objective federated learning. The first experiment is conducted to compare the performance of federated learning using sparse neural network models with that using fully connected networks. The second experiment employs the widely used NSGA-II to achieve a set of Pareto optimal solutions which should be validated in both IID and non-IID environments.

\subsection{Experimental settings}
In this section, we introduce some experimental settings in our case study. The settings include the following main parts: 1) Neural network models we used in the experiment and their original settings. 2) Parameters settings and data partition methods in federated learning. 3) Parameters of NSGA-II. 4) SET parameters for sparse connection.

We select two popular neural network models: the multilayer perceptron neural network (MLP) and the convolutional neural network (CNN), both trained and tested on a benchmark data set MNIST \cite{lecun1998gradient}. In optimizing both MLPs and CNNs, the mini-batch SGD algorithm has a learning rate of $0.1$ and the batch size is 50. Our original MLP contains two hidden layers, each having 200 nodes (199,210 parameters in total) and uses the ReLu function as the activation function, as used in \cite{mcmahan2016communication}. The CNN model has two $ 3 \times 3\ $ kernel filters (the first with 32 channels and the second with 64 channels) followed by a $ 2 \times 2\ $ Max pooling layer, a 128 fully connected layer and finally a 10 class softmax output layer (1,625,866 parameters in total). These can be seen as the \emph{\textbf{standard}} neural network structures in our experiments.

The total number of clients $K$ and a fraction of clients $C$ are set to be 100 and 1 in federated learning, meaning that we use $100 \times 1$ clients on each communication rounds. For each local client training, the mini-batch size $B$ and training epochs $E$ are 50 and 5, respectively. There are two ways of splitting MNIST dataset in our case study. One is IID, where the data is randomly shuffled into 100 clients with 600 samples per client, and the other is non-IID, where we sort the whole MNIST dataset by the labelled class, then divide it evenly into 200 fragments, and randomly allocate two fragments to each client with only two classes.

The population size of NSGA-II is set to be 20 due to limited computational resources. The evolutionary optimization is run for 20 generations on the IID dataset and 50 generations on the non-IID dataset, because we are more interested in the learning performance on the non-IID data. The parameters of crossover and mutation operators are empiriclly set as follows. We apply one-point crossover with a probability of 0.9 and bit-flip mutation with a probability of 0.1 to the binary chromosome, and the SBX with a probability of 0.9 probability and $ {n_c} = 2 $, and the polynomial mutation with a probability of 0.1 and $ {n_m} = 20\ $ \cite{kumar1995real} for the real-coded chromosome. In addition, the communication round required for fitness evaluations in NSGA-II is set to be 5 on the IID data and 10 on the non-IID data, because the global model trained on IID data needs less communication rounds to converge. Of course, evaluating fitness functions with a larger number of communication rounds can achieve more accurate fitness evaluations, but we are not allowed to do so, given very limited computation resources.

There are two SET parameters $ \varepsilon $ and $ \xi $ controlling the sparsity level of our models in federated learning. A pair of empirical values $ \varepsilon  = 20\ $and $ \xi  = 0.3\ $are implemented in \cite{mocanu2018scalable} for both MLPs and CNNs, which are also adopted in this work. In principle, these two parameters can also be binary coded and real coded, respectively, in genotypes for evolutionary optimization.

\subsection{Influence of the neural network sparsity on the performance}
In the first part of our experiment, we propose different settings of the SET parameters for both MLPs and CNNs to examine the influence of different sparsity levels on global model test accuracy and discuss model convergence properties on both the server and the client in federated learning.

Three different $ \varepsilon \ $values (100, 50, 20) and two different $ \xi \ $values (0, 0.3) are selected for both MLPs and CNNs with the standard structures (the original fully connected structure introduced above), which derives the standard federated models with different sparseness. Note that the modified SET algorithm applied on the FedAvg algorithm removes a $ \xi $ fraction of the least important weights at the last iteration of each local training epoch before being uploaded to the server. The parameters of the global model on the server are aggregated by calculating the weighted average of the uploaded models as done in the standard federated learning.

In addition, both MLPs and CNNs are tested on the IID and non-IID data and we run the modified SET FedAvg algorithm for 500 communication rounds for the MLPs and 200 communication rounds for CNNs. The reason for setting a smaller number of communication rounds for CNNs is that CNNs in federated learning are easier to converge but consume more time for a single communication round compared to that for MLPs. The results are shown in Fig. 6 and Fig. 7 for MLPs and CNNs, respectively.

\begin{figure}[!t]
\begin{minipage}[t]{1\linewidth}
\centering
\subfigure[MLP IID Global]{
\begin{minipage}[b]{0.46\textwidth}
\includegraphics[width=1\textwidth]{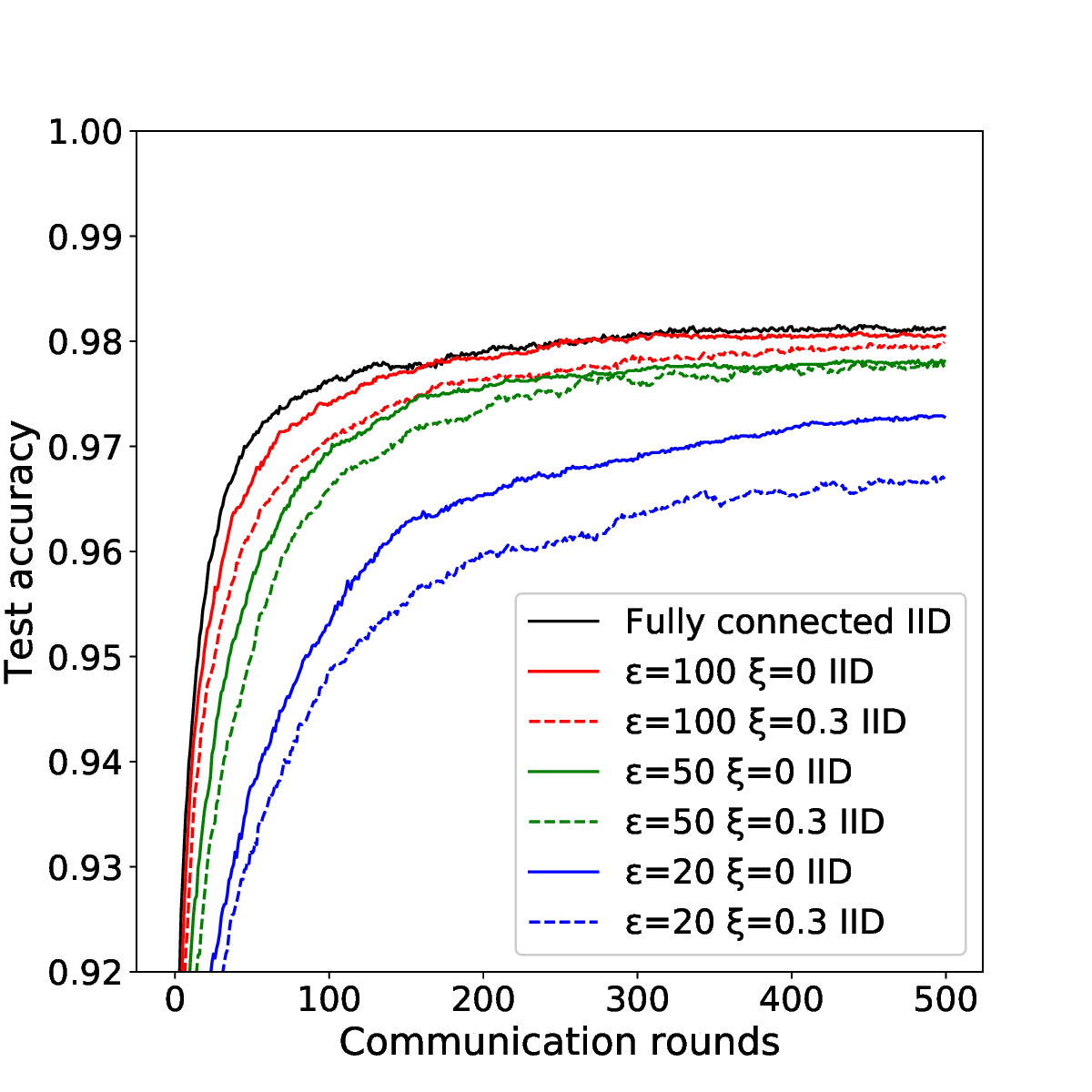}
\end{minipage}
}
\centering
\subfigure[MLP IID ClientAvg]{
\begin{minipage}[b]{0.46\textwidth}
\includegraphics[width=1\textwidth]{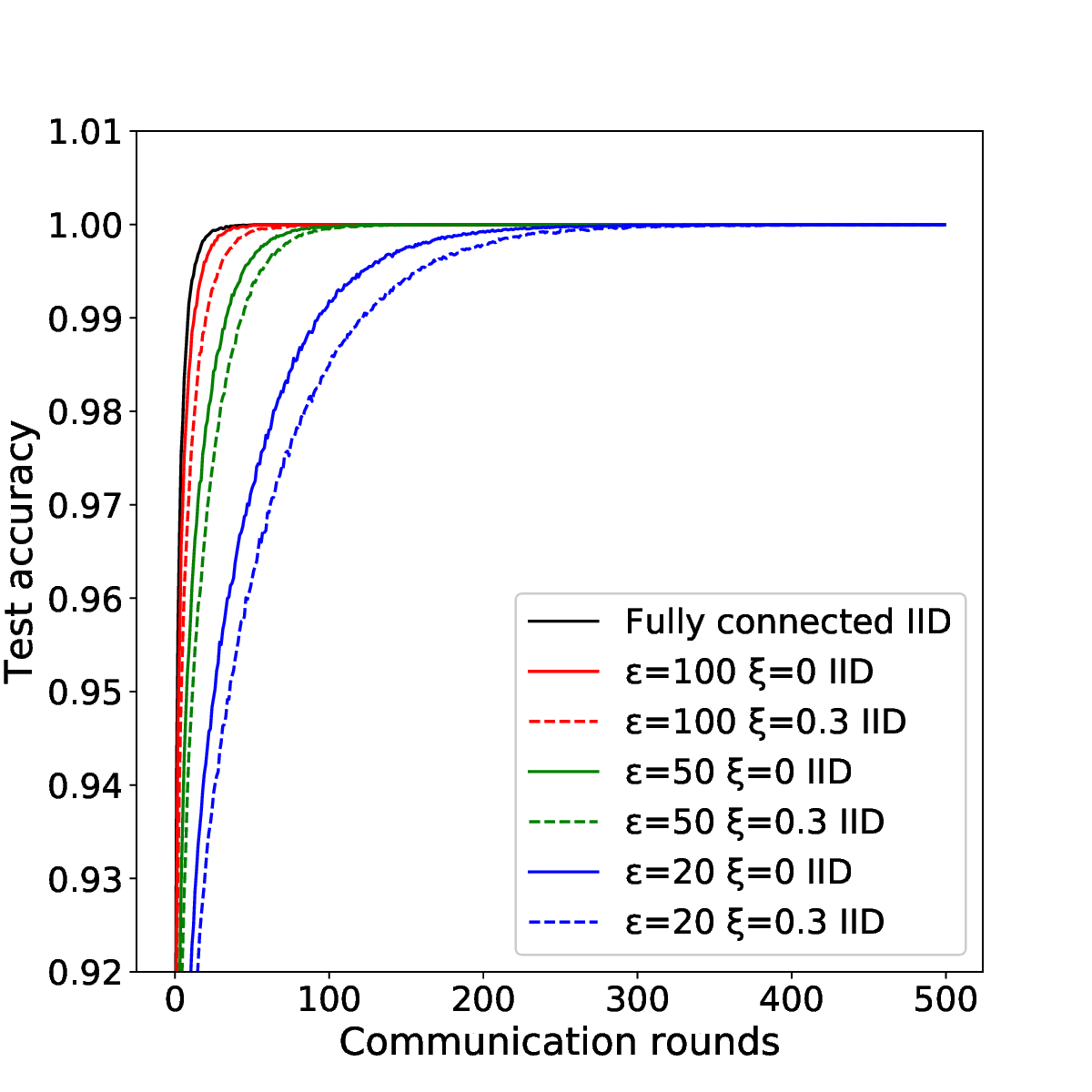}
\end{minipage}
} \\
\centering
\subfigure[MLP non-IID Global]{
\begin{minipage}[b]{0.46\textwidth}
\includegraphics[width=1\textwidth]{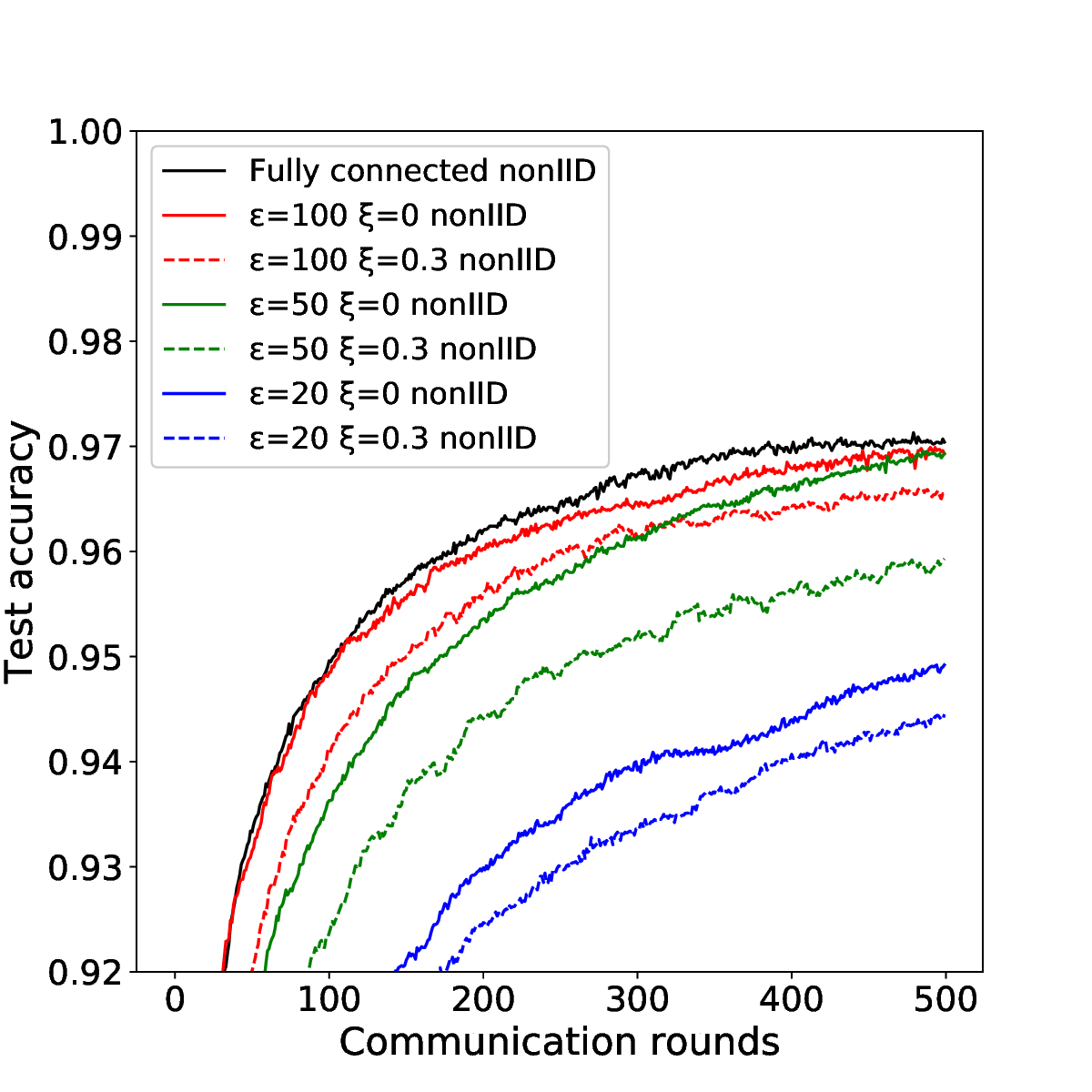}
\end{minipage}
}
\centering
\subfigure[MLP non-IID ClientAvg]{
\begin{minipage}[b]{0.46\textwidth}
\includegraphics[width=1\textwidth]{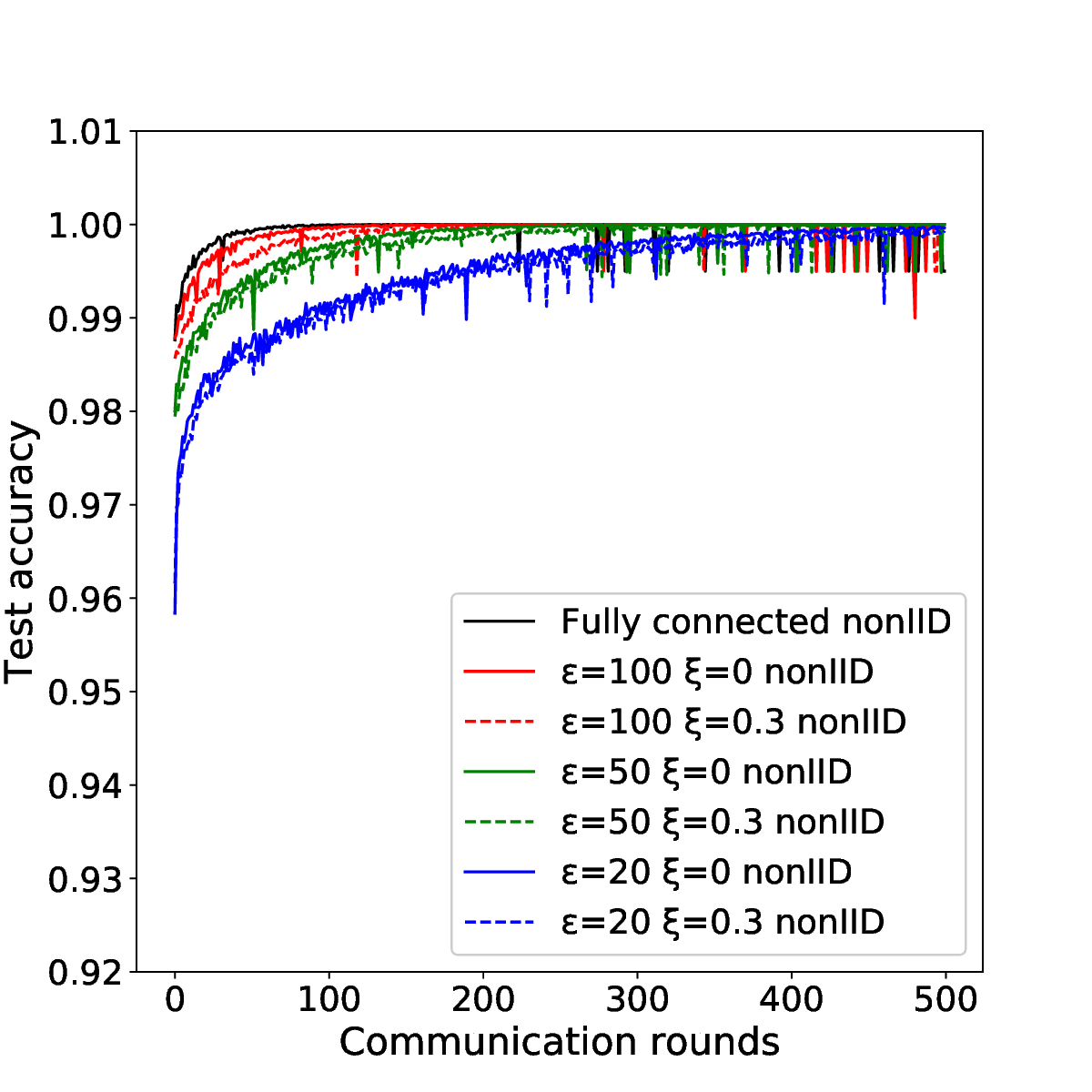}
\end{minipage}
} \\
\caption{The global model test accuracies and average client accuracies of MLPs on the IID and non-IID datasets. We select SET parameters $ \varepsilon \ $ and $ \xi \ $to be (100,0), (100,0.3), (50,0), (50.0.3), (20,0), (20,0.3), and total communications rounds to be 500.}
\label{Fig_5}
\end{minipage}
\end{figure}

\begin{figure}[!t]
\begin{minipage}[t]{1\linewidth}
\centering
\subfigure[CNN IID Global]{
\begin{minipage}[b]{0.46\textwidth}
\includegraphics[width=1\textwidth]{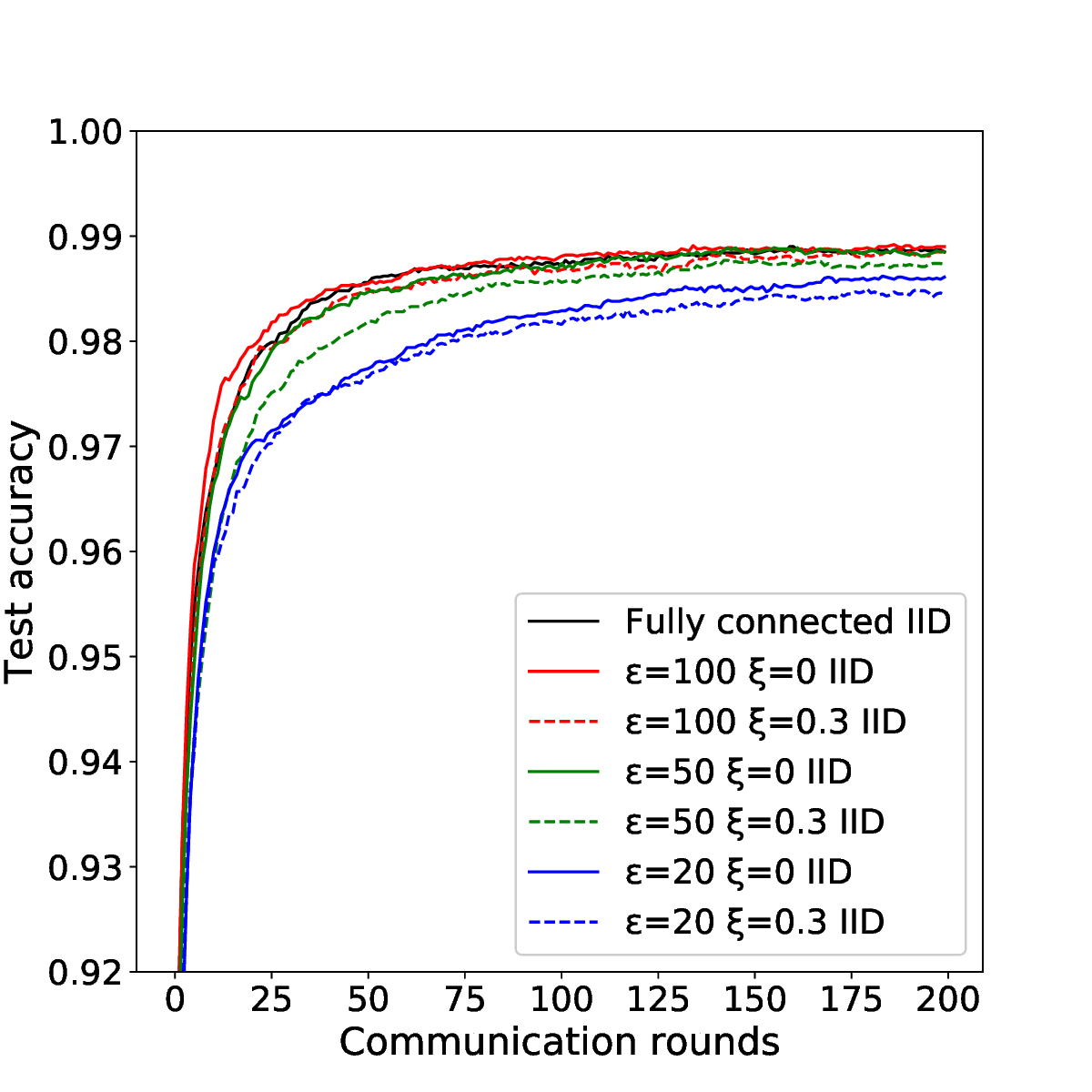}
\end{minipage}
}
\centering
\subfigure[CNN IID ClientAvg]{
\begin{minipage}[b]{0.46\textwidth}
\includegraphics[width=1\textwidth]{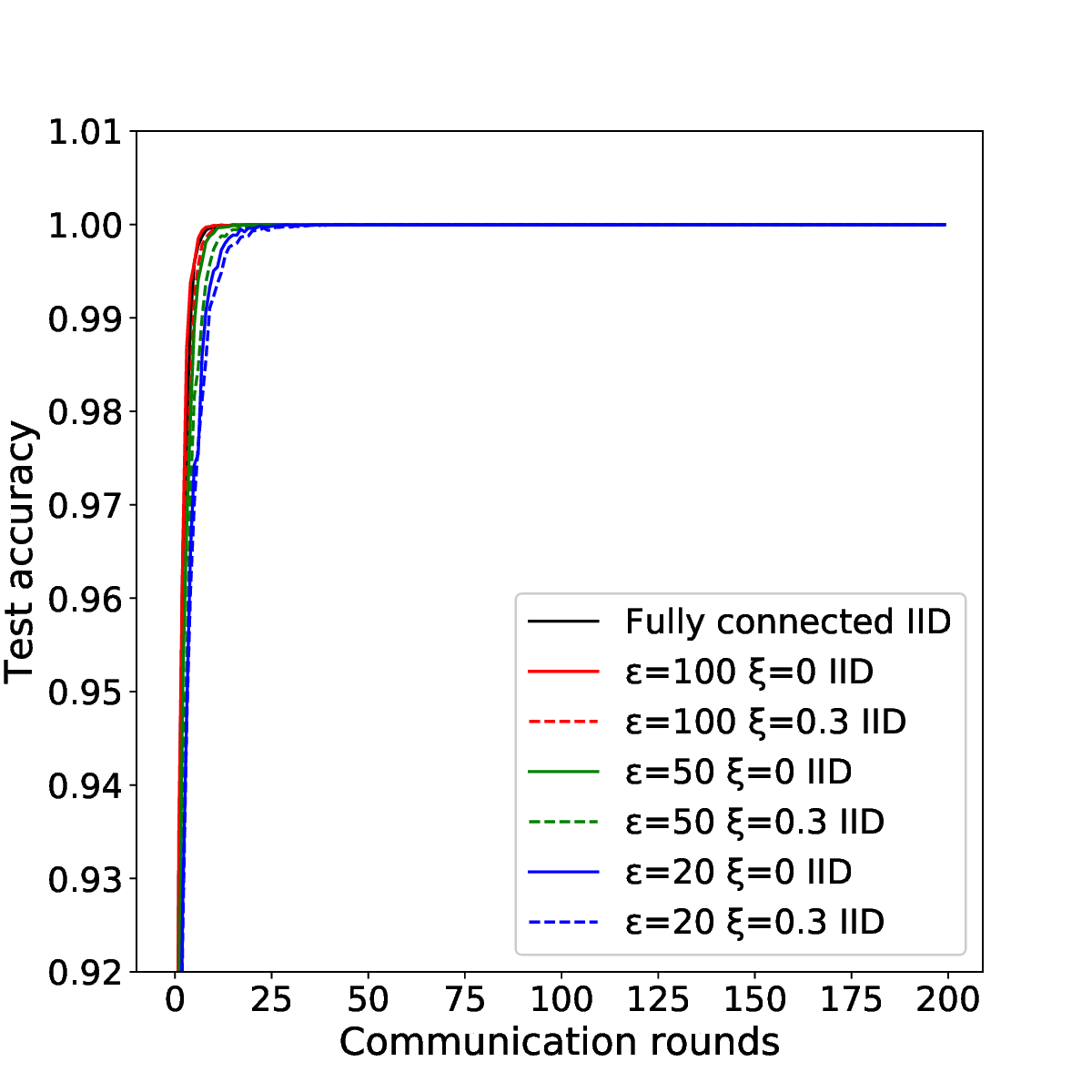}
\end{minipage}
} \\
\centering
\subfigure[CNN non-IID Global]{
\begin{minipage}[b]{0.46\textwidth}
\includegraphics[width=1\textwidth]{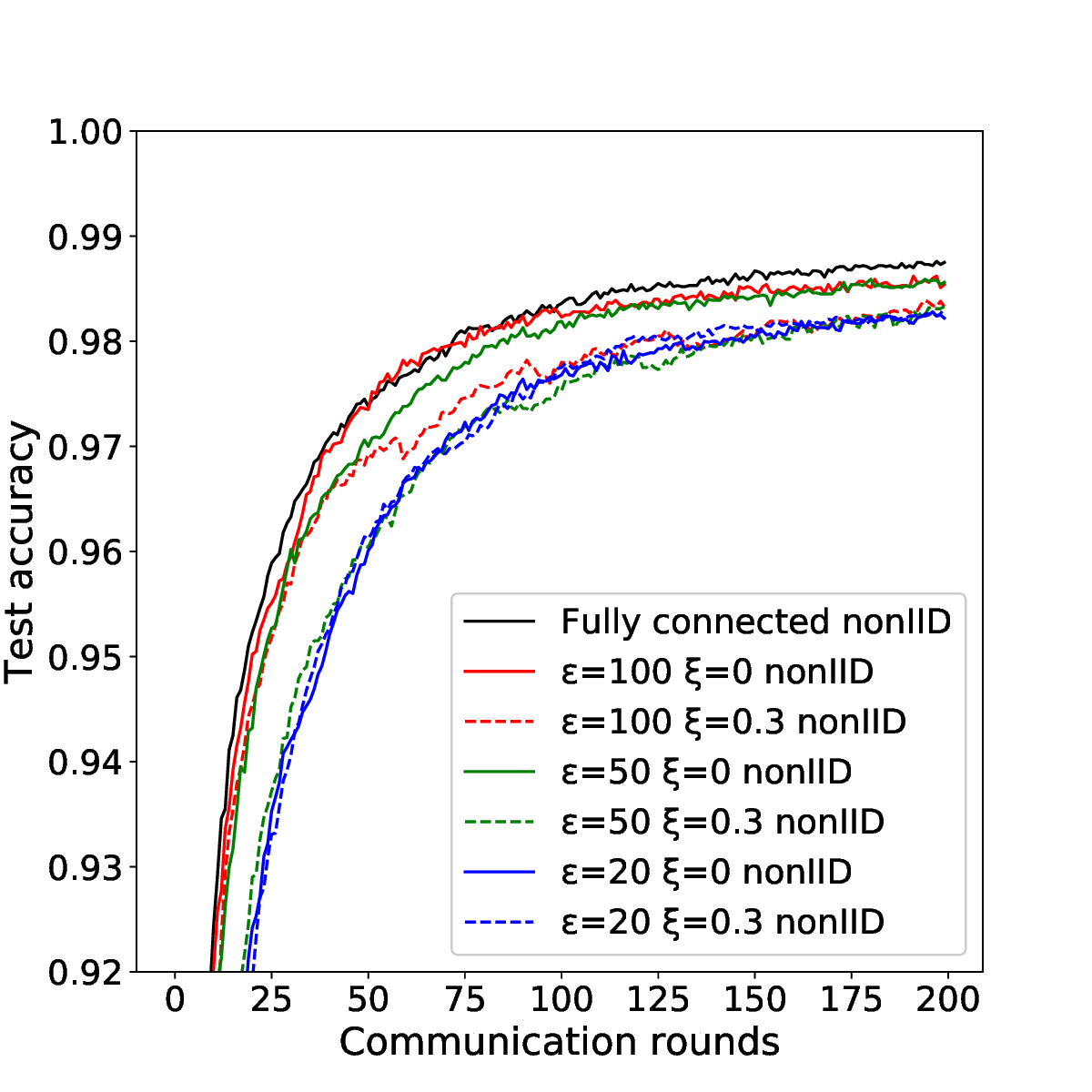}
\end{minipage}
}
\centering
\subfigure[CNN non-IID ClientAvg]{
\begin{minipage}[b]{0.46\textwidth}
\includegraphics[width=1\textwidth]{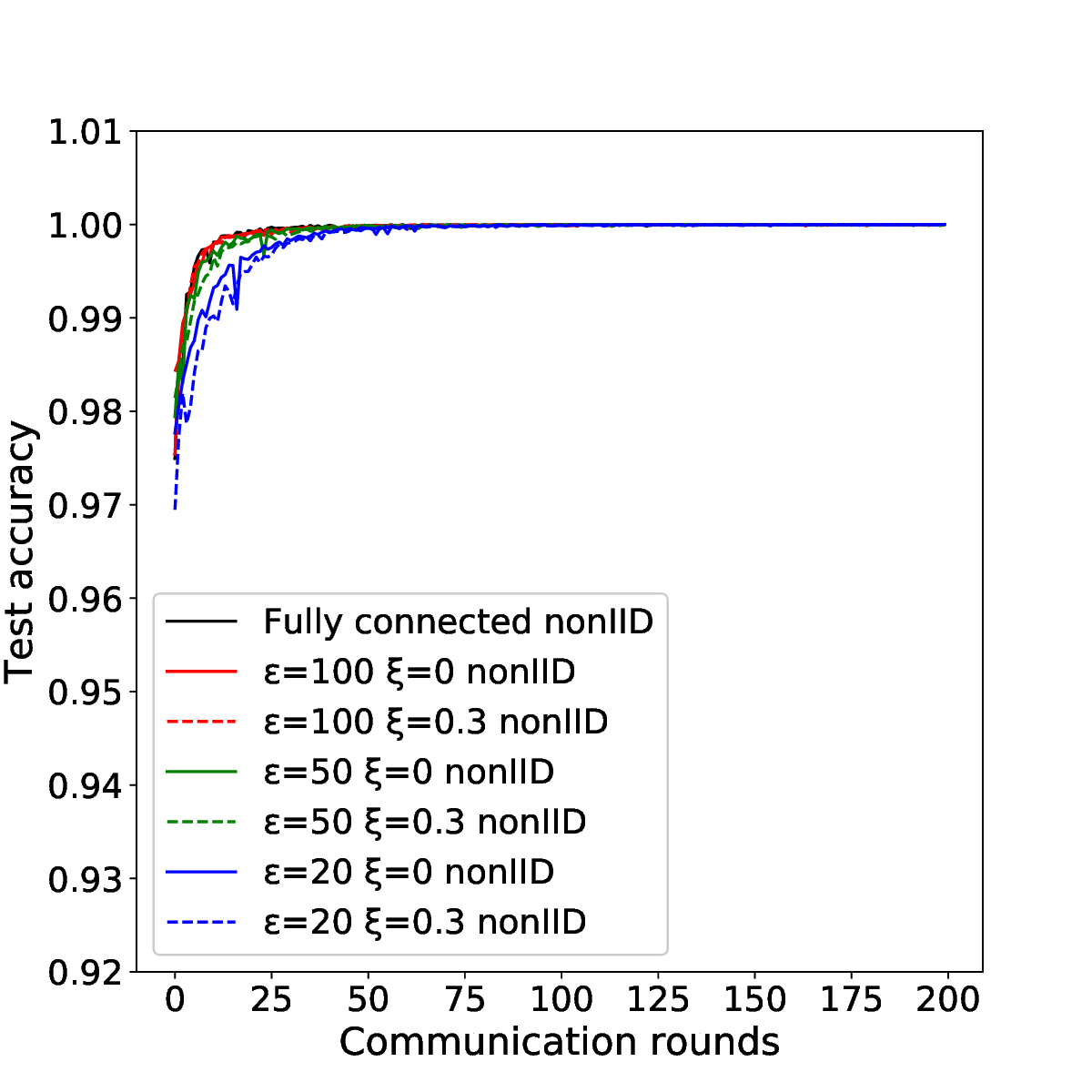}
\end{minipage}
} \\
\caption{The global model test accuracies and averaged client accuracies of CNNs on the IID and non-IID data sets. We select SET parameters $ \varepsilon \ $ and $ \xi \ $to be (100,0), (100,0.3), (50,0), (50.0.3), (20,0), (20,0.3), and total communications rounds to be 200.}
\label{Fig_6}
\end{minipage}
\end{figure}

We discuss at first the convergence properties of the shared models on the server and the clients when learning the IID and non-IID data.
The convergence performance on the clients is assessed through calculating the average training accuracy over all clients. The average training accuracy reaches nearly 100\% within only a few rounds on the non-IID data and also becomes higher than 95$ \% \ $ within the first 25 to 50 rounds of communication on the IID data, refer to Figs. 6 (b)(d) and Figs. 7 (b)(d). By contrast, learning converges much slower on the server, in particular on the non-IID data, as shown in Figs. 6(a)(c) and Figs. 7(a)(c). This indicates that learning on the server becomes more challenging, in particular on non-IID data.

To take a closer look at the learning behavior on the server, we compare the global test accuracies on the server as the sparsity level of the neural network models varies. An observation that can be made from the results in Figs. 6(a)(c) and Figs. 7(a)(c) is that reducing the network connectivity may lead to a degradation of the global test accuracies on both IID and non-IID dataset. However, the test accuracy enhances as $ \xi $ decreases, i.e., when less 'least important' weights are removed from neural network models on each client before uploading them to the server. For instance, a global test accuracy of 96.93$ \%  $ has been achieved when $ \varepsilon  = 50, \xi  = 0 $ in the SET algorithm that result in 72051 connections on average, as shown in Fig. 6(c). This accuracy is higher than 96.54$ \%  $ when the SET parameters $ \varepsilon  = 100, \xi  = 0.3 $ that result in 87300 connections on average at the $500$-th round. This implies that removing a larger fraction of weights is detrimental to the learning performance.

Nevertheless, it is clearly seen that there is a trade-off between the global test accuracy and average model complexity of the local models. The experimental results of the fully connected neural network model and mostly sparsely connected neural network model found by the proposed algorithm (whose SET parameters are $ \varepsilon  = 20,\xi  = 0.3 $) are listed in Table \uppercase\expandafter{\romannumeral1}. We can see that the global test accuracies of the sparsely connected MLPs (having about only 10$ \% \ $of the total number connections in the fully connected models) is only about 2$ \% \ $lower than that of the fully connected one on both IID and non-IID datasets. The global test accuracy of the sparse CNN, which has only about 12$ \% \ $of the total number of connections of the fully connected CNN, is only 0.45$ \%  $ worse than the fully connected CNN. Moreover, it should be pointed out that test accuracies of both MLPs and CNNs deteriorate more quickly on the non-IID data than on the IID data as the sparsity level of the network increases.

Overall, the global model test accuracy on the server tends to decline when we tune the SET parameters to rise the sparseness of the shared neural network model in our experiment. In other words, using the modified SET FedAvg algorithm only cannot maximize the global learning accuracy and minimize the communication costs at the same time.

\begin{table}[]
\caption{Global test accuracies and the number of average connections}
\setlength{\tabcolsep}{1mm}{
\begin{tabular}{llllll}
\hline
\multicolumn{2}{l}{Local data distributions}                   & \multicolumn{2}{l}{IID} & \multicolumn{2}{l}{non-IID} \\ \cline{3-6}
\multicolumn{2}{l}{}                                           & Accuracy  & Connections & Accuracy    & Connections   \\ \hline
\multicolumn{1}{l}{\multirow{2}{*}{Fully connected}}    & MLP & 98.13\%   & 199,210     & 97.04\%     & 199,210       \\
\multicolumn{1}{l}{}                                    & CNN & 98.85\%   & 1,625,866   & 98.75\%     & 1,625,866     \\
\multicolumn{1}{l}{\multirow{2}{*}{Sparsely connected}} & MLP & 96.69\%   & 19,360      & 94.45\%     & 18,785        \\
\multicolumn{1}{l}{}                                    & CNN & 98.44\%   & 185,407     & 98.32\%     & 184,543       \\ \hline
\end{tabular}}
\end{table}


\subsection{Evolved federated learning models}
In the second part of our empirical studies, we employ NSGA-II to achieve a set of Pareto optimal neural network models that balance a trade-off between global learning performance and communication costs in federated learning. Both IID and non-IID datasets will be used in multi-objective evolutionary optimization of federated learning. It is also interesting to investigate if the structure of the neural network models optimized on IID datasets still work on non-IID datasets, and vice versa.

Evolving deep neural network structures based on the modified SET FedAvg algorithm is computationally highly intensive. For example, one run of evolutionary optimization of CNNs with a population size of 20 for 50 generations took us more than one week on a computer with GTX 1080Ti GPU and i7-8th 8700 CPU, preventing us from running the evolutionary optimization for a large number of generations. In order to monitor the convergence of the multi-objective optimization, the hypervolumes calculated based on the non-dominated solution set in the population over the generations \cite{beume2007sms} in evolving MLP and CNN on non-IID datasets are plotted in Fig. 8. From the figure, we can see that the hypervolumes of both runs increase at the beginning and start fluctuating from around the $20$-th generation onward. These results imply that approximately 20 generations are needed for federated learning to converge on non-IID datasets used in this work.

\begin{figure}[!t]
\begin{minipage}[t]{1\linewidth}
\centering
\subfigure[Hypervolume for MLP]{
\begin{minipage}[b]{0.46\textwidth}
\includegraphics[width=1\textwidth]{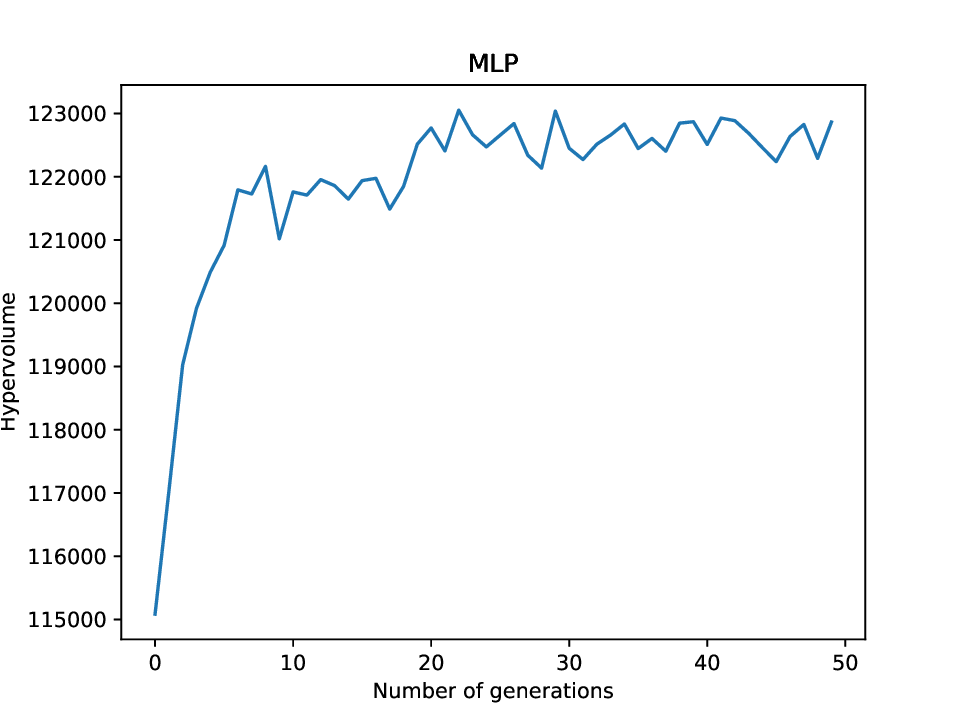}
\end{minipage}
}
\centering
\subfigure[Hypervolume for CNN]{
\begin{minipage}[b]{0.46\textwidth}
\includegraphics[width=1\textwidth]{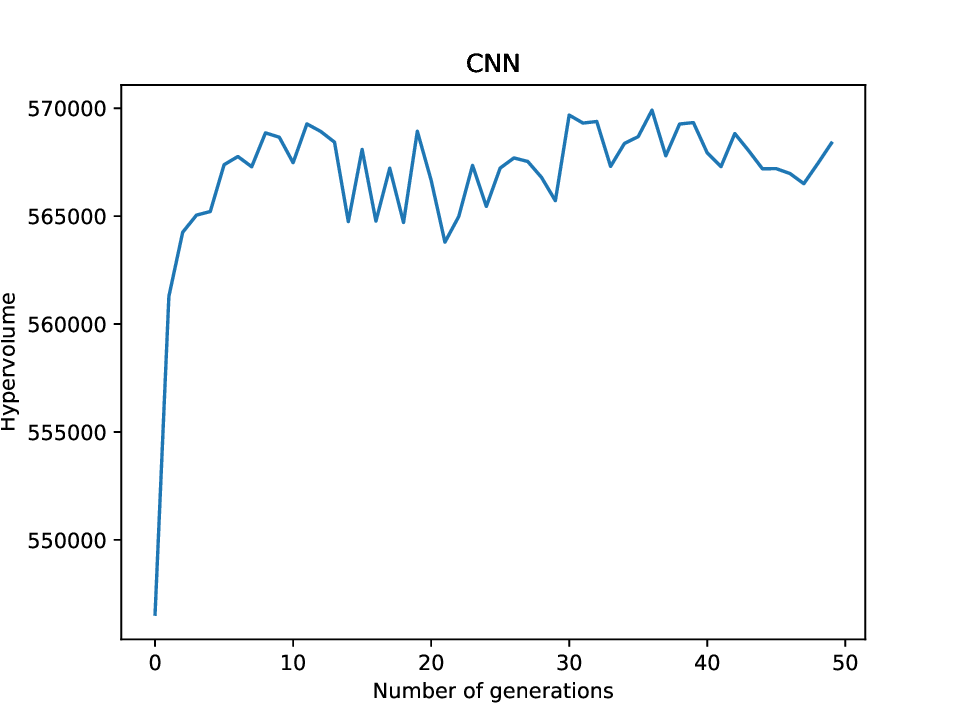}
\end{minipage}
} \\
\caption{Change of hypervolume over the generations in training MLP and CNN on non-IID datasets.}
\label{Fig_7}
\end{minipage}
\end{figure}

The total communication rounds for each population is set to be 5 for IID datasets and 10 for non-IID datasets, respectively, before the objective values are calculated. Of course, setting a large communication rounds may achieve more accurate evaluations of the objectives, which is unfortunately prohibitive given limited computation resources.

We set the maximum number of hidden layers of MLPs to be 4 and the maximum number of neurons per layer is 256. For CNNs, we set the maximum number of convolutional layers to be 3, the maximum number of kernel channels to be 64, and the maximum number of fully connected layers to be 3, and the maximum neurons in the convolutional layers to be 256. The kernel size is either 3 or 5, which is also evolved.

The range of the learning rate is between 0.01 and 0.3 for both MLPs and CNNs, because too large values may harm the global convergence in federated learning.

Recall that the SET parameters $ \varepsilon \ $and $ \xi  $ are binary coded and real coded, respectively. The maximum value of $ \varepsilon  $ is set to 128, and $ \xi \ $ranges from 0.01 to 0.55. A summary of the experimental settings is given in Table \ref{T_Settings}.

\begin{table}[]
\centering
\caption{Experimental settings for multi-objective optimization of federated learning\label{T_Settings}}
\setlength{\tabcolsep}{1mm}{
\begin{tabular}{lllll}
\hline
Genotypes      & MLP IID   & MLP nonIID & \multicolumn{1}{l}{CNN IID} & \multicolumn{1}{l}{CNN nonIID} \\ \hline
Populations   & 20        & 20         & 20                          & 20                             \\
Generations   & 20        & 50         & 20                          & 50                             \\
Learning rate & 0.01-0.3  & 0.01-0.3   & 0.01-0.3                    & 0.01-0.3                       \\
Hidden layers & 1-4       & 1-4        & /                           & /                              \\
Hidden neurons   & 1-256         & 1-256          & /                           & /                              \\
Conv layers   & /         & /          & 1-3                           & 1-3                              \\
Kernel channels   & /         & /          & 1-64                           & 1-64                              \\
Fully connected layers     & /         & /          & 1-3                           & 1-3                              \\
Fully connected neurons       & /         & /          & 1-256                           & 1-256                              \\
Kernel sizes   & /         & /          & 3 or 5                      & 3 or 5                         \\
$\varepsilon$ sizes  & 1-128         & 1-128          & 1-128                           & 1-128                              \\
$ \xi  $ sizes         & 0.01-0.55 & 0.01-0.55  & 0.01-0.55                   & 0.01-0.55                      \\ \hline
\end{tabular}}
\end{table}

The final non-dominated MLP and CNN solutions optimized on the IID and non-IID datasets are presented in Fig. 9 and Fig. 10, respectively, where each point represents one solution corresponding to a particular structure of the neural network model in federated learning. However, not all non-dominated solutions are of interest, since some of them have very large test errors, even if they have very simple model structures with very limited average local model connections. In this work, we select two types of non-dominated solutions, namely those solutions with a very low global test error, and those solutions near the knee point of the frontier, as suggsted in \cite{jin2008pareto}.

\begin{figure}[!t]
\begin{minipage}[t]{1\linewidth}
\centering
\subfigure[Evolved Pareto frontier of MLPs trained on IID datasets]{
\begin{minipage}[b]{0.46\textwidth}
\includegraphics[width=1\textwidth]{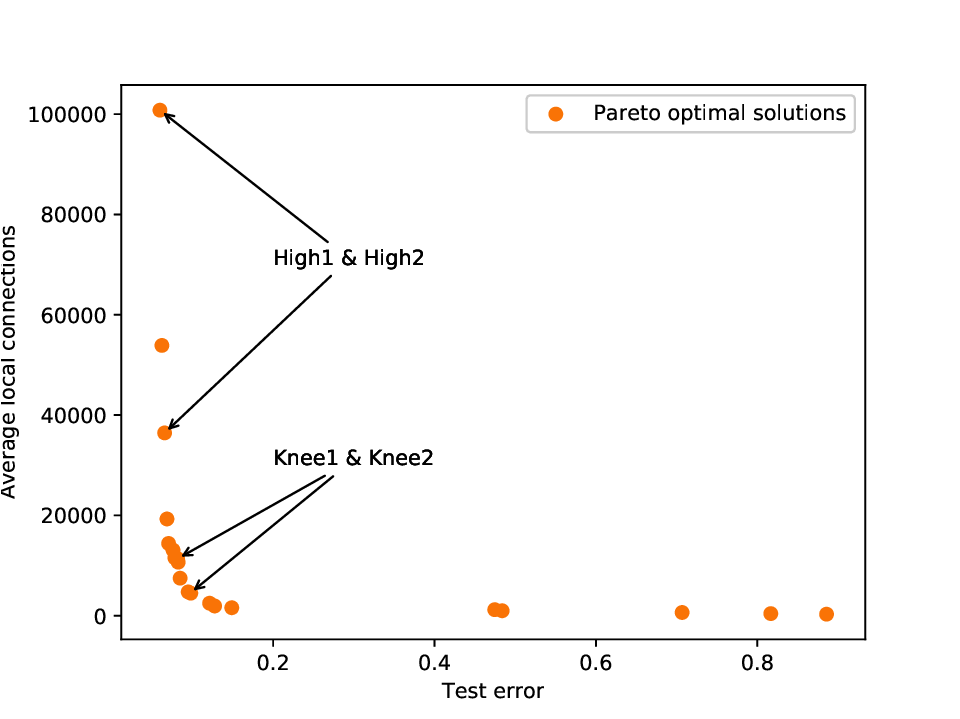}
\end{minipage}
}
\centering
\subfigure[Evolved Pareto frontier of MLPs trained on non-IID datasets]{
\begin{minipage}[b]{0.46\textwidth}
\includegraphics[width=1\textwidth]{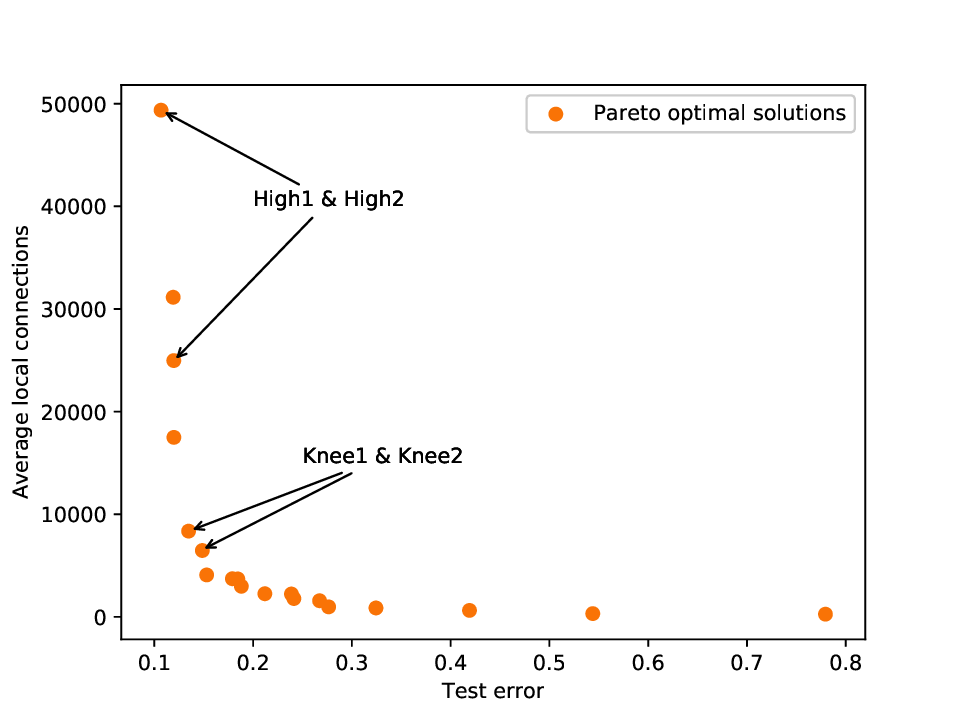}
\end{minipage}
} \\
\caption{Pareto frontier of MLPs, of which four solutions, High1, High2, Knee1, Knee2 are selected for validation.}
\label{Fig_8}
\end{minipage}
\end{figure}
\begin{figure}[!t]
\begin{minipage}[t]{1\linewidth}
\centering
\subfigure[Evolved Pareto frontier of CNNs trained on IID datasets]{
\begin{minipage}[b]{0.46\textwidth}
\includegraphics[width=1\textwidth]{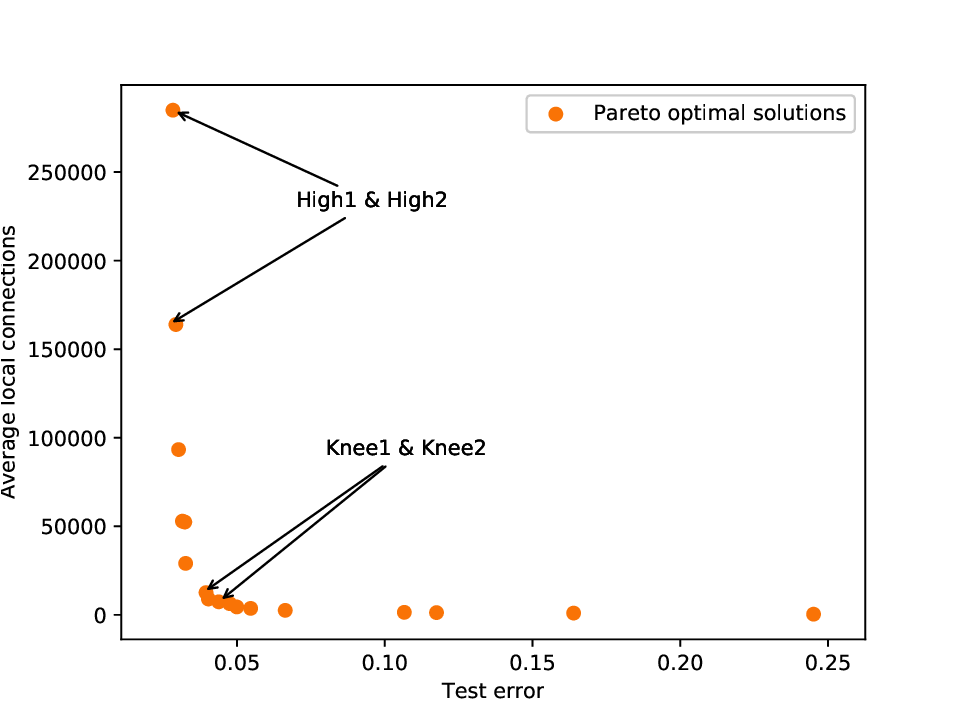}
\end{minipage}
}
\centering
\subfigure[Evolved Pareto frontier of CNNs trained on non-IID datasets]{
\begin{minipage}[b]{0.46\textwidth}
\includegraphics[width=1\textwidth]{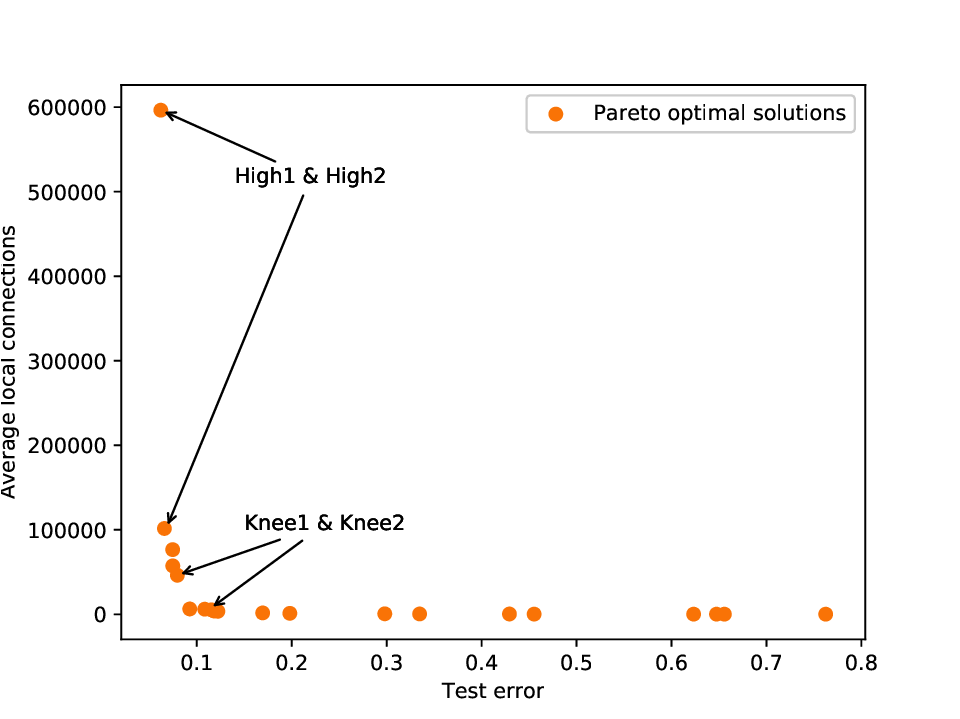}
\end{minipage}
} \\
\caption{Pareto frontier of CNNs, of which High1, High2, Knee1, and Knee2 are selected for validation.}
\label{Fig_9}
\end{minipage}
\end{figure}

We choose two high-accuracy Pareto solutions (High1 and High2) and two solutions around the knee point (Knee1 and Knee2) of both MLPs and CNNs (refer to Fig. 9 and Fig. 10) for further performance verification and compare their performance with the fully connected MLPs and CNNs. Recall that only 5 and 10 communication rounds are used over IID data and non-
IID data, respectively, for fitness evaluations in the evolutionary optimization. For a fair comparison, however, the number communication rounds is increased to 500 for MLPs and 200 for CNNs, as set in the original federated learning. All validation results are listed in Tables \ref{T_Vali_MLP_IID}, \ref{T_Vali_MLP_nonIID}, \ref{T_Vali_CNN_IID}, and \ref{T_Vali_CNN_nonIID}, and the global test accuracies of the selected solutions are also presented in Fig. 11 and Fig. 12.

\begin{table}[]
\centering
\caption{Hyper-parameters of High1, High2, Knee1, and Knee2 for \textbf{\emph{MLPs}} evolved on \textbf{\emph{IID}} data and their validation results \label{T_Vali_MLP_IID}}
\setlength{\tabcolsep}{1.2mm}{
\begin{tabular}{llllll}
\hline
Parameters           & Knee1   & Knee2   & High1   & High2   & Standard \\ \hline
Hidden layer1        & 10      & 15      & 152     & 73      & 200      \\
Hidden layer2        & 27      & 123     & 49      & 22      & 200      \\
$\varepsilon$              & 28      & 60      & 121     & 48      & /        \\
$ \xi  $                 & 0.3969  & 0.2021  & 0.1314  & 0.1214  & /        \\
Learning rate $\eta $      & 0.2591  & 0.3     & 0.2951  & 0.283   & 0.1      \\
Test accuracy IID    & 94.24\% & 96.84\% & 98.16\% & 97.74\% & 98.13\%  \\
Connections IID      & 4,374   & 10,815  & 91,933  & 32,929  & 199,210  \\
Test accuracy nonIID & 90.77\% & 93.77\% & 97.42\% & 96.82\% & 97.04\%  \\
Connections nonIID   & 4,026   & 10,206  & 91,527  & 33,594  & 199,210  \\ \hline
\end{tabular}}
\end{table}

\begin{table}[]
\centering
\caption{Hyper-parameters of High1, High2, Knee1, and Knee2 for \textbf{\emph{MLPs}} evolved on \emph{\textbf{non-IID}} data and their validation results \label{T_Vali_MLP_nonIID}}
\setlength{\tabcolsep}{1.2mm}{
\begin{tabular}{llllll}
\hline
Parameters           & Knee1   & Knee2   & High1   & High2   & Standard \\ \hline
Hidden layer1        & 49      & 53      & 86      & 109     & 200      \\
Hidden layer2        & /       & /       & /       & /       & 200      \\
$\varepsilon$              & 10      & 8       & 66      & 34      & /        \\
$ \xi  $                 & 0.1106  & 0.0764  & 0.1106  & 0.1566  & /        \\
Learning rate $\eta $      & 0.3     & 0.2961  & 0.3     & 0.3     & 0.1      \\
Test accuracy IID    & 96.78\% & 96.41\% & 97.82\% & 97.68\% & 98.13\%  \\
Connections IID      & 7,749   & 5,621   & 45,329  & 22,210  & 199,210  \\
Test accuracy nonIID & 94.85\% & 94.88\% & 97.32\% & 96.21\% & 97.04\%  \\
Connections nonIID   & 8,086   & 6,143   & 45,530  & 24,055  & 199,210  \\ \hline
\end{tabular}}
\end{table}

\begin{table}[]
\centering
\caption{Hyper-parameters of High1, High2, Knee1, ad Knee2 for \textbf{\emph{CNNs}} evolved on \textbf{\emph{IID}} data and their validation results \label{T_Vali_CNN_IID}}
\setlength{\tabcolsep}{1.2mm}{
\begin{tabular}{llllll}
\hline
Parameters           & Knee1   & Knee2   & High1   & High2   & Standard  \\ \hline
Conv layer1          & 34      & 6       & 25      & 18      & 32        \\
Conv layer2          & 6       & 6       & 38      & 20      & 64        \\
Fully connected layer1     & 11      & 9       & 38      & 102     & 128       \\
Fully connected layer2     & /       & /       & /       & /       & /         \\
Kernel size          & 5       & 5       & 5       & 5       & 3         \\
$\varepsilon$              & 24      & 39      & 121     & 41      & /         \\
$ \xi  $                  & 0.4702  & 0.3901  & 0.0685  & 0.0625  & /         \\
Learning rate $\eta $       & 0.2094  & 0.1576  & 0.2279  & 0.1888  & 0.1       \\
Test accuracy IID    & 98.51\% & 98.19\% & 99.07\% & 98.96\% & 98.85\%   \\
Connections IID      & 12,360  & 7,127   & 268,150 & 158,340 & 1,625,866 \\
Test accuracy nonIID & 11.35\% & 97.21\% & 11.35\% & 98.79\% & 98.75\%   \\
Connections nonIID   & 6,071   & 6,804   & 24,853  & 157,511 & 1,625,866 \\ \hline
\end{tabular}}
\end{table}

\begin{table}[]
\centering
\caption{Hyper-parameters of High1, High2, Knee1, and Knee2 for \textbf{\emph{CNNs}} evolved on \textbf{\emph{non-IID}} data and their validation results \label{T_Vali_CNN_nonIID}}
\setlength{\tabcolsep}{1.2mm}{
\begin{tabular}{llllll}
\hline
Parameters           & Knee1   & Knee2   & High1   & High2   & Standard  \\ \hline
Conv layer1          & 17      & 5       & 53      & 33      & 32        \\
Conv layer2          & /       & /       & /       & /       & 64        \\
Fully connected layer1            & 29      & 21      & 208     & 31      & 128       \\
Fully connected layer2            & /       & /       & /       & /       & /         \\
Kernel size          & 5       & 5       & 5       & 5       & 3         \\
$\varepsilon$              & 18      & 8       & 66      & 20      & /         \\
$ \xi  $                 & 0.1451  & 0.1892  & 0.0786  & 0.1354  & /         \\
Learning rate $\eta $        & 0.2519  & 0.2388  & 0.2776  & 0.2503  & 0.1       \\
Test accuracy IID    & 98.84\% & 98.15\% & 99.06\% & 98.93\% & 98.85\%   \\
Connections IID      & 48949   & 6262    & 622090  & 107224  & 1,625,866 \\
Test accuracy nonIID & 97.92\% & 97.7\%  & 98.52\% & 98.46\% & 98.75\%   \\
Connections nonIID   & 39457   & 6804    & 553402  & 90081   & 1,625,866 \\ \hline
\end{tabular}}
\end{table}

\begin{figure}[!t]
\begin{minipage}[t]{1\linewidth}
\centering
\subfigure[Solutions evolved on IID data and validated on IID data]{
\begin{minipage}[b]{0.46\textwidth}
\includegraphics[width=1\textwidth]{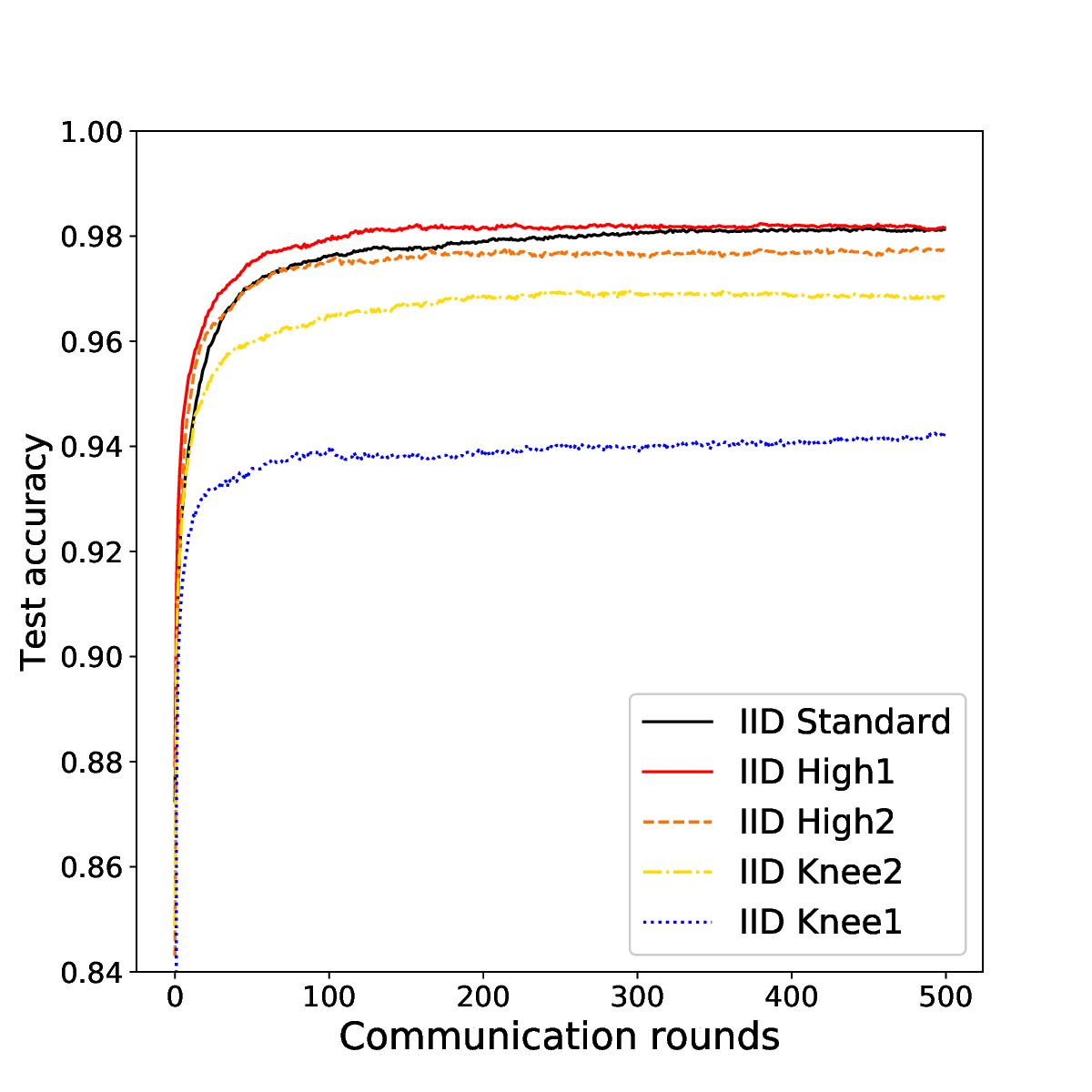}
\end{minipage}
}
\centering
\subfigure[Solutions evolved on IID data and validated on non-IID data]{
\begin{minipage}[b]{0.46\textwidth}
\includegraphics[width=1\textwidth]{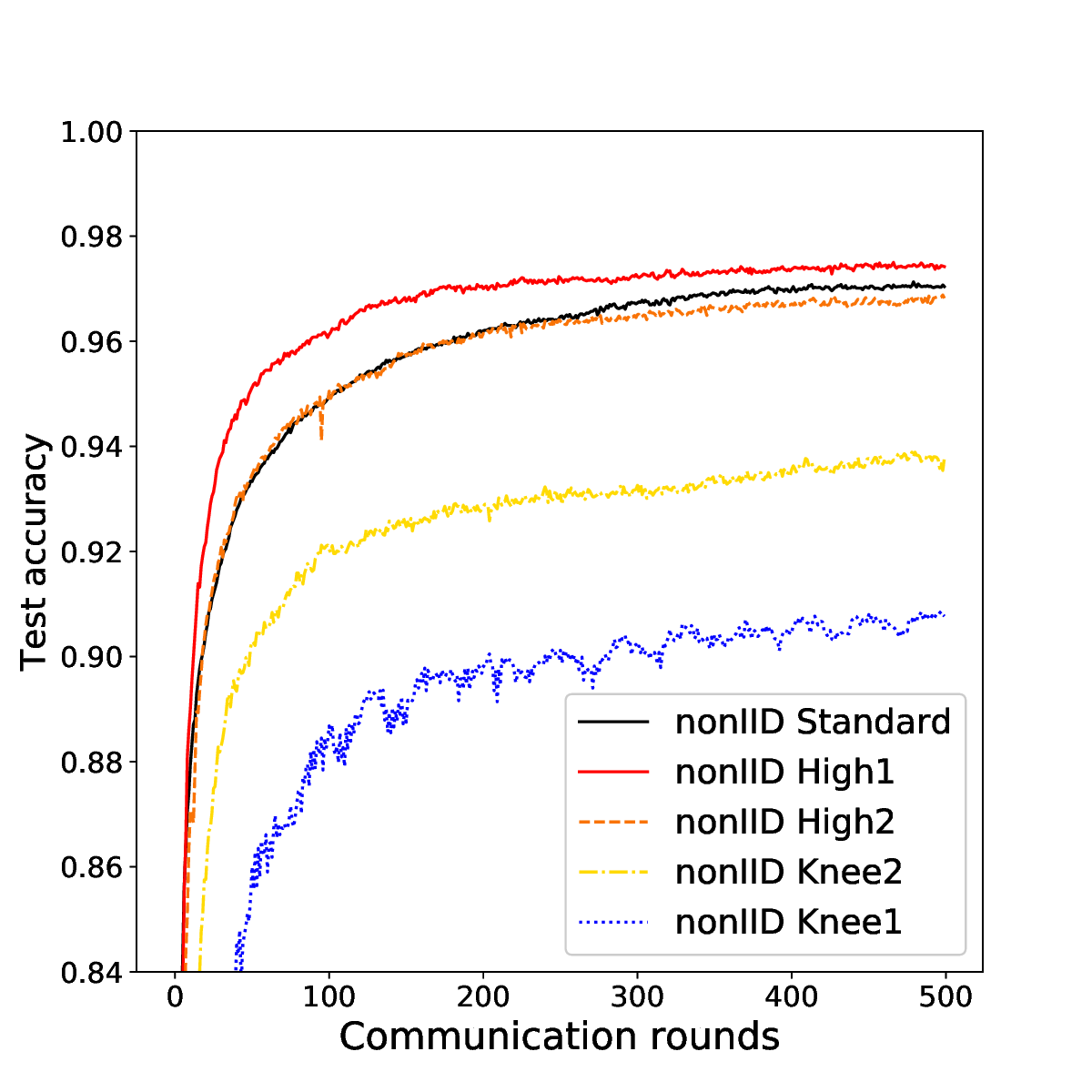}
\end{minipage}
} \\
\centering
\subfigure[Solutions evolved on non-IID data and validated on IID data]{
\begin{minipage}[b]{0.46\textwidth}
\includegraphics[width=1\textwidth]{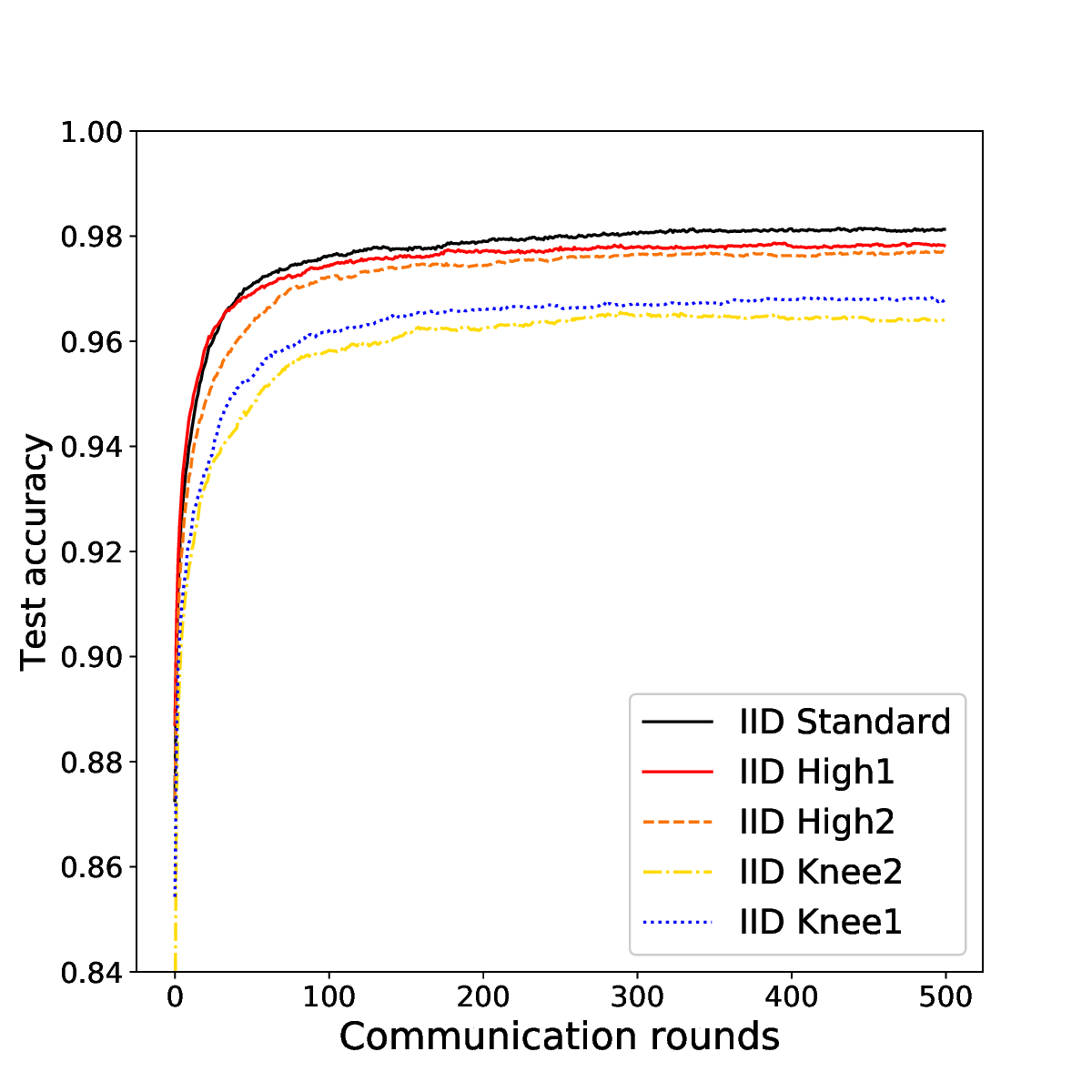}
\end{minipage}
}
\centering
\subfigure[Solutions evolved on non-IID data and validated on non-IID data]{
\begin{minipage}[b]{0.46\textwidth}
\includegraphics[width=1\textwidth]{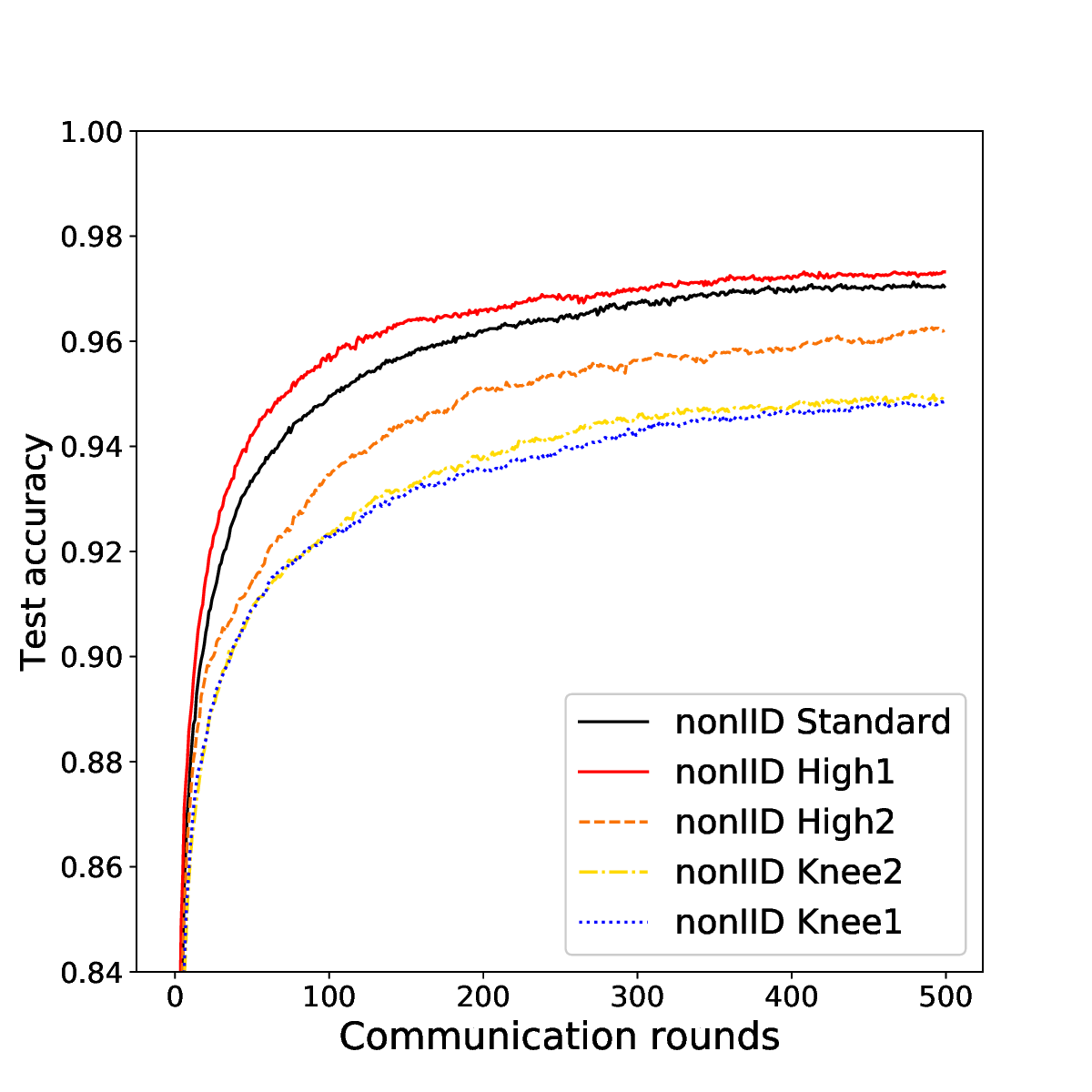}
\end{minipage}
} \\
\caption{The global test accuracies of the selected Pareto optimal MLPs validated on both IID and non-IID data. The test accuracies of the fully connected MLP are also plotted in the figure for comparison.}
\label{Fig_MLP_Val}
\end{minipage}
\end{figure}

\begin{figure}[!t]
\begin{minipage}[t]{1\linewidth}
\centering
\subfigure[Solutions evolved on IID data and validated on IID data]{
\begin{minipage}[b]{0.46\textwidth}
\includegraphics[width=1\textwidth]{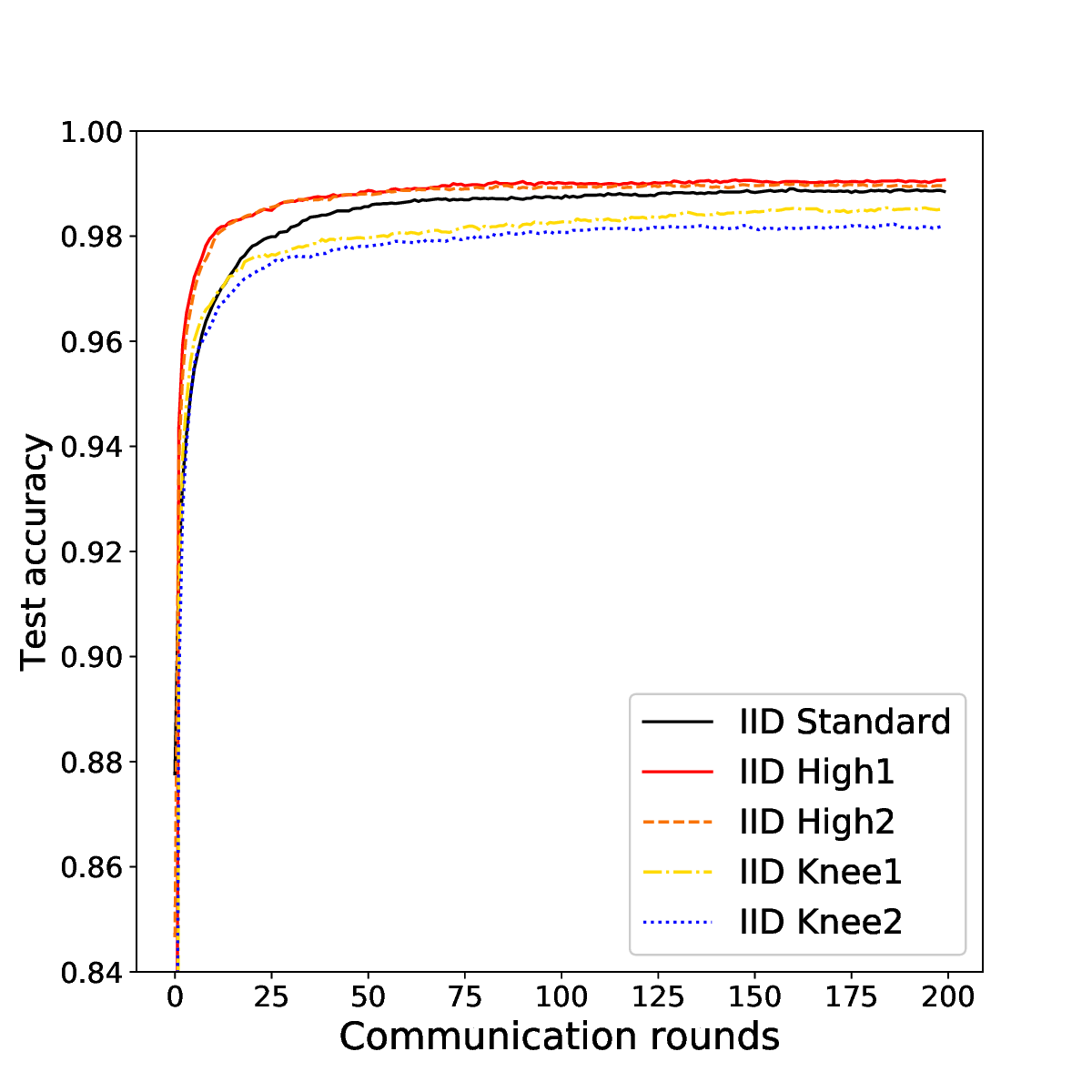}
\end{minipage}
}
\centering
\subfigure[Solutions evolved on IID data and validated non-IID data]{
\begin{minipage}[b]{0.46\textwidth}
\includegraphics[width=1\textwidth]{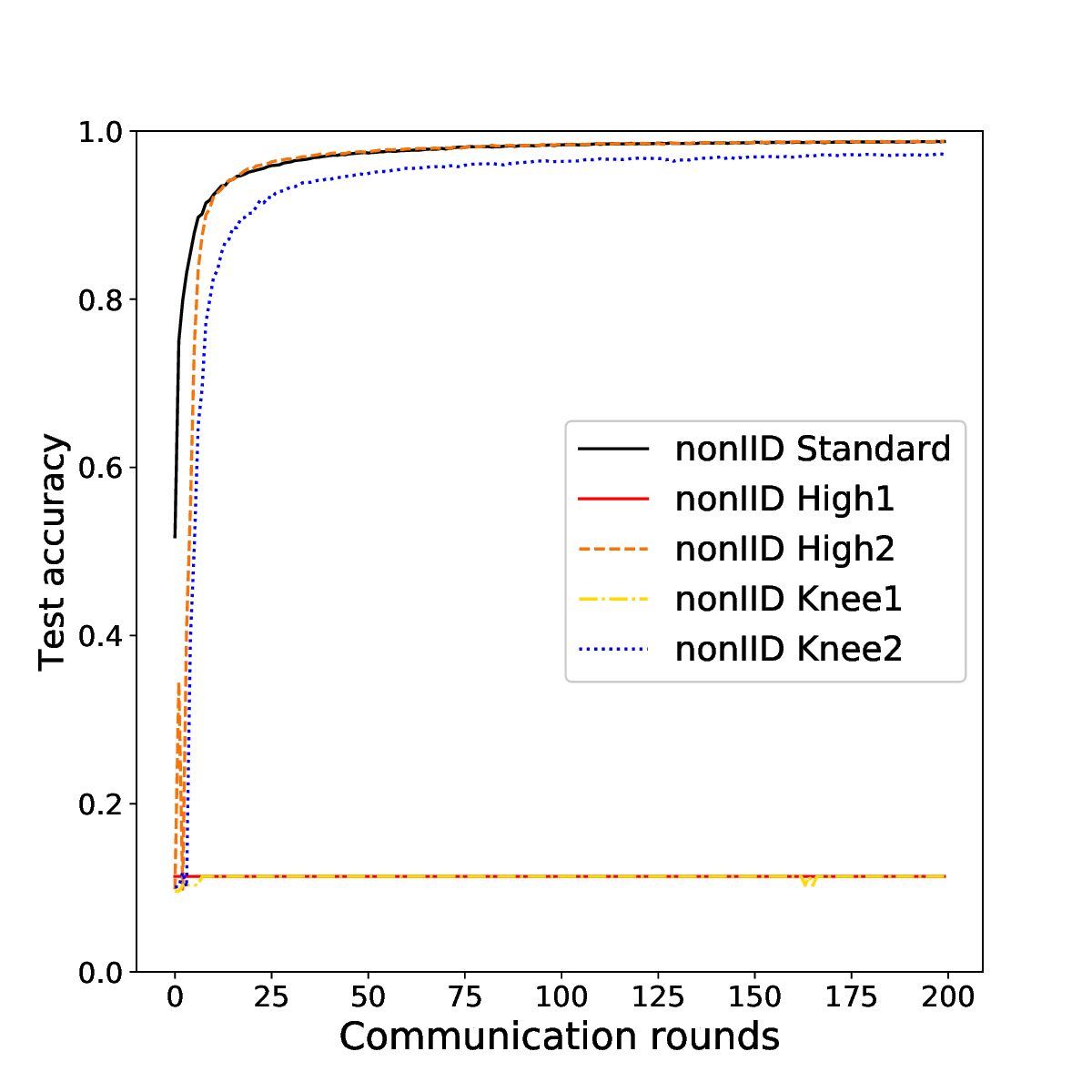}
\end{minipage}
} \\
\centering
\subfigure[Solutions evolved on non-IID data and validated on IID data]{
\begin{minipage}[b]{0.46\textwidth}
\includegraphics[width=1\textwidth]{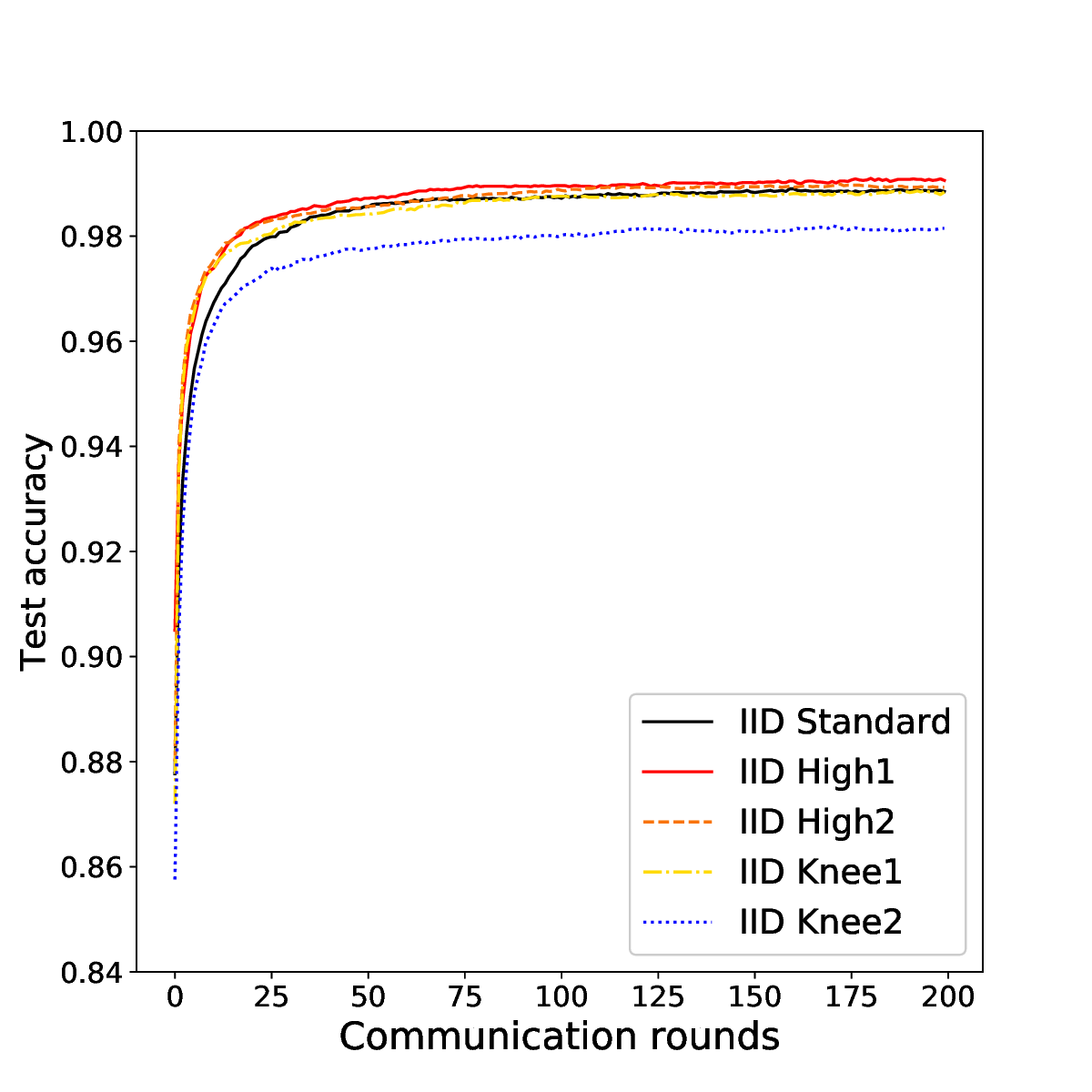}
\end{minipage}
}
\centering
\subfigure[Solutions evolved ob non-IID data and validated on non-IID data]{
\begin{minipage}[b]{0.46\textwidth}
\includegraphics[width=1\textwidth]{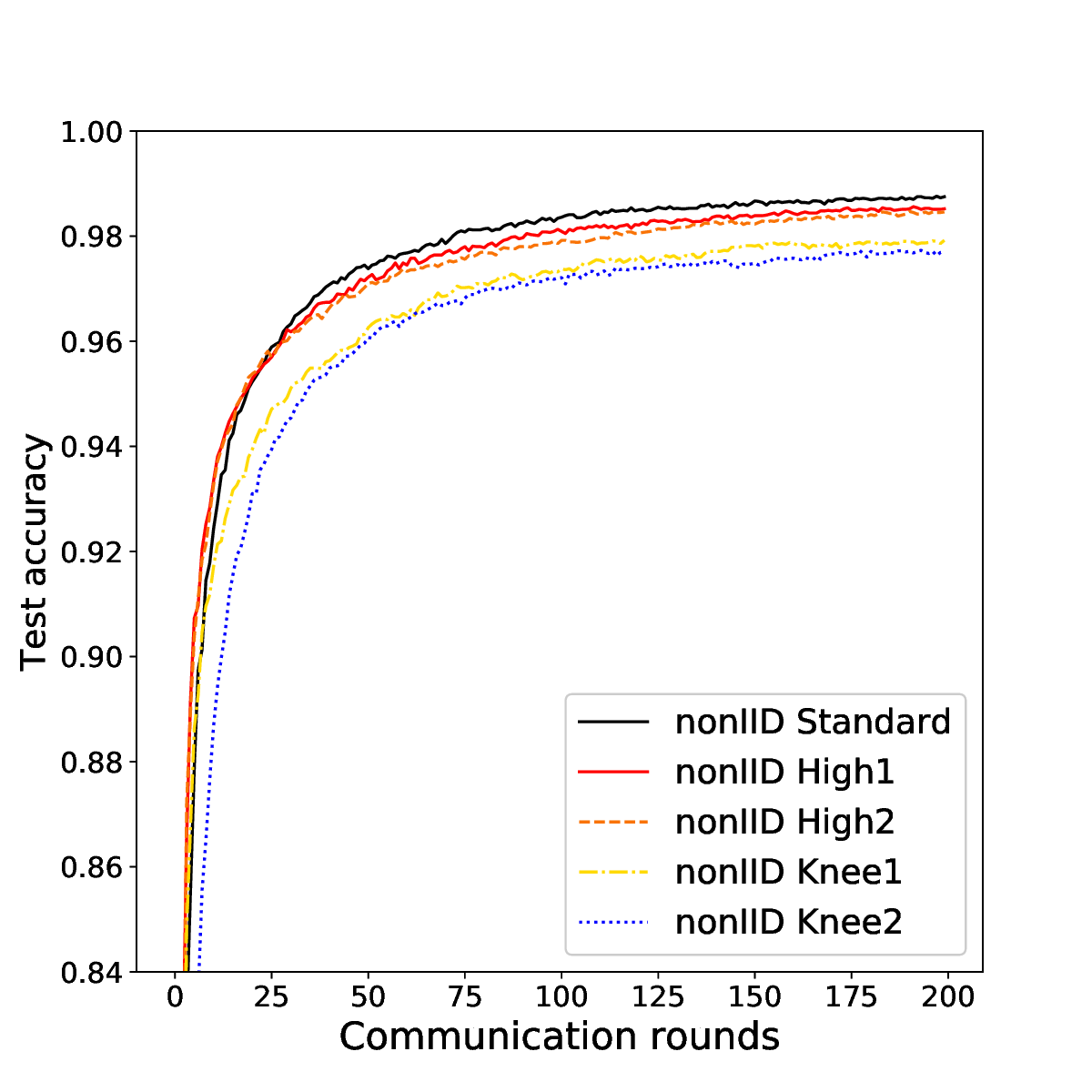}
\end{minipage}
} \\
\caption{The global test accuracies of the selected Pareto optimal CNNs validated on both IID and non-IID data. The test accuracies of the fully connected CNN are also plotted for comparison.}
\label{Fig_CNN_Val}
\end{minipage}
\end{figure}

From the results presented in Figs. \ref{Fig_MLP_Val} and \ref{Fig_CNN_Val}, we can make the following observations on the four selected Pareto optimal MLP models evolved on the IID data.
\begin{itemize}
\item Solution High1 of MLP has global test accuracies of 98.16$ \% \ $and 97.42$ \% \ $on IID and non-IID datasets, both of which are better than that of the fully connected MLP. In addition, this evolved model has only on average 91,933 and 91,527 connections on IID data and non-IID data, respectively, which is approximately 46$ \%  $ of 199,210 connections the fully connected network has.

\item Solution High2 has a lower test accuracy of 0.39$ \% \ $ on IID data and 0.22$ \% \ $on non-IID data but it has only 16.5$ \% \ $ of connections compared to the fully connected MLP.

\item Knee1 and Knee2 have test accuracies of 96.84$ \% \ $and 94.24$ \%  $, respectively, on IID datasets. Note, however, that but their performance becomes much worse on non-IID data and the test accuracies decrease to only 93.77$ \% \ $and 90.77$ \%  $. This means that knee solutions evolved on IID data may not be suited for non-IID data.
\end{itemize}

Similar observations can be made on the two high-accuracy Pareto optimal CNNs, High1 and High2, on IID data and their test accuracies are 99.07$ \% \ $and 98.96$ \%  $, respectively, both of which are higher than that of the fully connected CNN. The two knee solutions also have acceptable global test accuracies on IID data with a much smaller number of connections. However, it is surprising to see that both Knee1 and High1 fail to converge on the non-IID data, even if they both converge very well on the IID data, meaning that the Pareto optimal CNNs generated on IID data may completely break down on non-IID data. Note that the high-accuracy solution High2 also converge well on non-IID data and has a test accuracy of 98.79$ \%  $, which is 0.04$ \% \ $higher than the fully connected model, but has only around 10$ \% \ $of connections of the fully connected CNN.

Based on the above validation results, we recommend to select the following solutions from the Pareto frontier for final implementation. The high-accuracy MLP solution, High1 with two hidden layers should be selected. This model has 152 and 49 neurons in the first and second hidden layers, respectively. SET parameters of the network are $ \varepsilon  = 121 $ and $ \xi  = 0.1314 $, and the learning rate is 0.2951. It has better global test accuracies than the fully connected model on both IID and non-IID data, while has around 46$ \% \ $of the connections of the standard fully connected network. By contrast, the high-accuracy CNN solution High2 with two $ 5 \times 5 $ convolutional layers should be selected. The first and second layers of this network have 18 and 20 filters, respectively. The fully connected layer has 102 nodes and the SET parameters are $ \varepsilon  = 41 $ and $ \xi  = 0.0625 $ and the learning rate is 0.1888. The network has better global test accuracies than the standard fully connected model on both IID and non-IID data, but has only about 9.7$ \% \ $of the connections of the standard fully connected networks. In addition, one knee point CNN solution Knee1 has one $ 5 \times 5 $ convolutional layer, 17 kernel filters, 29 nodes in the fully connected layer, whose SET parameters are $ \varepsilon  = 18 $ and $ \xi  = 0.1415 $ and learning rate is 0.2591, has a similar global test accuracy on IID data, and 0.8$ \%  $ worse than the fully connected one on non-IID data. This network has only about 3$ \% \ $of the connections of the standard fully connected model. Thus, Knee1 of the CNN model is also recommendable.

The following observations can be made from our experimental results:
\begin{itemize}
\item The proposed algorithm can achieve a set of Pareto optimal solutions, from which we can select multiple solutions based on different preferences for different learning tasks.
\item The solutions we selected for comparison, either the knee points or solutions with a high accuracy, can significantly reduce the number of parameters in the global model to be transferred between the server and clients without seriously deteriorating the model performance in federated learning. Actually, the proposed algorithm has also found two solutions that have higher test accuracies than the original settings on both IID and non-IID datasets, one being an MLP that has 46$\%$ of the connections in the standard MLP and the other being a CNN that has only 9.7$\%$ the connections in the fully connected one. By significantly reducing the global model size transferred between devices, the proposed multiobjective evolutionary algorithm can effectively enhance the communication efficiency at a rate of at least 50$\%$ in training a neural network model in federated learning.
\item The global model structures in federated learning evolved from non-IID data are more robust than those evolved from the IID data. Specifically, our solutions in the optimal Pareto frontier evolved from the IID dataset may not fit very well when the data becomes non-IID. As seen from our previous experimental results, some of our solutions cannot converge at all. On the contrary, solutions evolved from the non-IID dataset also performs well on the IID dataset. And this implies that it is harder for federated learning on non-IID data to converge than the traditional distributed learning only on IID data.
\item The model structures evolved on IID datasets are usually deeper than that evolved on non-IID datasets. In addition, the proposed algorithm allows different clients have different model sizes, which is computationally more efficient and enables more reduction in communication cost.
\end{itemize}

\section{Conclusions and future work}
This work proposes a multi-objective federated learning to simultaneously maximize the learning performance and minimize the communication cost using a multi-objective evolutionary algorithm. To improve the scalability in evolving large neural networks, a modified SET method is suggested to indirectly encode the connectivity of the neural network. Our experimental results demonstrate that the modified SET algorithm can effectively reduce the number of the connections of neural networks by encoding only two hyper parameters. Selected solutions from the non-dominated frontier obtained by the multi-objective algorithm confirm that the proposed algorithm is able to generate neural network models for federated learning that exhibit better global learning accuracy and have much fewer connections, thereby dramatically reducing the communication cost without deteriorating the learning performance on both IID and non-IID datasets.

A lot of work remain to be done in federated learning. For instance, both the modified SET FedAvg algorithm and the FedAvg algorithm do not work very well on complicated datasets like non-IID CIFAR-10. Although the proposed algorithm is applicable to deep networks in principle, the network models studied in this work are fairly simple. Thus, it is of great interest to investigate the performance of the proposed algorithm in optimizing deep neural networks having dozens of hidden layers. In addition, it is still unclear if missing data caused by package loss in communications between clients and the server will significantly affect the performance of federated learning. Finally, adversarial attacks \cite{bagdasaryan2018backdoor} on the parameters uploaded to the central server may directly damage the global model. Thus, preserving privacy while maintaining robustness in federated learning will be a very important research challenge.

\section{Acknowledgement}
We are grateful to Y. Zhao for sharing his code.


%

%
%
%
%
%

\ifCLASSOPTIONcaptionsoff
  \newpage
\fi


%



{\footnotesize\bibliography{reference/ref}
\bibliographystyle{ieeetr}}
%


%
%




\end{document}